\documentclass{article}

\usepackage[final]{NFFL2021}

\usepackage[utf8]{inputenc} 
\usepackage[T1]{fontenc}    
\usepackage{url}            
\usepackage{booktabs}       
\usepackage{amsfonts}       
\usepackage{nicefrac}       
\usepackage{microtype}      
\usepackage{xcolor}         

\usepackage{graphicx} 
\usepackage{amsmath}
\usepackage{algorithm}
\usepackage{algpseudocode}
\usepackage{multirow}
\usepackage{array}
\usepackage{adjustbox}
\usepackage[title]{appendix}

\title{FedRAD: Federated Robust Adaptive Distillation}

\author{%
  Stefán Páll Sturluson  \\
  Department of Computing \\
  Imperial College London, UK \\
  \texttt{stefan.sturluson20@imperial.ac.uk}
  
  \And
  Samuel Trew \\
  Department of Computing \\
  Imperial College London, UK \\
  \texttt{samuel.trew17@imperial.ac.uk}
    
  \And
  Luis Muñoz-González\\
  Department of Computing \\
  Imperial College London, UK \\
  \texttt{l.munoz-gonzalez@imperial.ac.uk}
  
  \And
  Matei Grama \\
  Department of Computing \\
  Imperial College London, UK \\
  \texttt{matei.grama17@imperial.ac.uk}
  
  \And
  Jonathan Passerat-Palmbach\\
  Department of Computing \\
  Imperial College London, UK \\
  \texttt{j.passerat-palmbach@imperial.ac.uk}
  
  \And
  Daniel Rueckert\\
  Department of Computing \\
  Imperial College London, UK \\
  \texttt{d.rueckert@imperial.ac.uk} 
  
  \And
  Amir Alansary\\
  Department of Computing \\
  Imperial College London, UK \\
  \texttt{a.alansary14@imperial.ac.uk} \\
}

\begin{document}

\maketitle

\begin{abstract}
The robustness of federated learning (FL) is vital for the distributed training of an accurate global model that is shared among large number of clients.
The collaborative learning framework by typically aggregating model updates is vulnerable to model poisoning attacks from adversarial clients.
Since the shared information between the global server and participants are only limited to model parameters, it is challenging to detect bad model updates.
Moreover, real-world datasets are usually heterogeneous and not independent and identically distributed (Non-IID) among participants, which makes the design of such robust FL pipeline more difficult.
In this work, we propose a novel robust aggregation method, \textbf{Fed}erated \textbf{R}obust \textbf{A}daptive \textbf{D}istillation (FedRAD), to detect adversaries and robustly aggregate local models based on properties of the median statistic, and then performing an adapted version of ensemble Knowledge Distillation.
We run extensive experiments to evaluate the proposed method against recently published works.
The results show that FedRAD outperforms all other aggregators in the presence of adversaries, as well as in heterogeneous data distributions.
\end{abstract}

\section{Introduction}

In recent years, deep learning models have shown themselves to be powerful prediction tools, often outperforming simpler models \cite{krizhevsky2012imagenet}. They have been applied in a wide variety of fields, such as healthcare \cite{jiang2017healthcare}, machine translation \cite{bahdanau2014machineTranslation}, and autonomous driving \cite{bewley2019driving}.
However, such methods are data hungry, and require large datasets to perform well \cite{obermeyer2016dataHungry}. 
In practice, data is usually expensive and time consuming to collect and annotate. Besides, privacy-sensitive data cannot be directly shared globally, as it violates the current laws stated by regulatory bodies such as the general data protection regulations (GDPR)\footnote{https://gdpr-info.eu/} and the health insurance portability and accountability act
of (HIPAA)\footnote{https://www.hhs.gov/hipaa/index.html}.


Federated Learning (FL) enables model training collaboratively in a decentralized manner without the need for data owners to hand over their data \cite{mcmahan2017communication}. 
A typical FL pipeline consists of two main steps: (1) training a copy of the global model locally on the client's private data, and (2) aggregating the local parameters into an updated global model. These two steps are repeated until converging to a final model.
Model aggregation is usually performed by averaging the model updates (FedAvg). However, some of the participants could be faulty or malicious by sharing bad parameters' updates, which can ruin the global model's performance and prevent it from converging \cite{yin2018byzantine}.

Recent works \cite{grama2020robust, yin2018byzantine, blanchard2017machine, munoz2019byzantine, hu2020fedmgda+} have been published to mitigate the aforementioned limitations of FedAvg in the presence of malicious users and noisy adversaries. However, these methods fail to achieve the same good performance on heterogeneous data \cite{zhao2018federatedNonIID}. 
In practice, data heterogeneity is very common because of the different geographical locations or tools used to acquire the data. 
For example, the frequency of species of animals in natural image datasets collected in different continents will vary a lot, leading to clients from different continents to have label-skewed datasets. 

In this paper, we propose a novel robust aggregation method that achieves state-of-the-art performance on heterogeneous client data, while still being able to detect adversaries or bad users and effectively defend against their attacks. The main contributions of this paper can be summarized as:
\begin{itemize}
    \item  A novel robust aggregator called Federated Robust Adaptive Knowledge Distillation (FedRAD), which utilizes a median-based scoring system along with knowledge distillation. 
    \item Extensive evaluations of different aggregation methods with different types of attacks and levels of data heterogeneity.
    \item FedRAD outperformed all other methods in the presence of adversaries, both in IID and non-IID settings.
    \item The code is open source and publicly available, which can accelerate significantly the research in robust FL.
\end{itemize}



\section{Background}

\textbf{Standard aggregation:}
Deep learning models are trained using stochastic gradient descent (SGD), where the parameters' updates are computed on a mini-batch sampled from a local dataset, and back-propagated through the model.
FedSGD \cite{shokri2017membership} is a federated leaning pipeline based on sampling a fraction of the gradients for every client, which are then sent to and averaged by a global server proportionally to the number of the samples on each client.
Federated Averaging (FedAvg) \cite{mcmahan2017communication} is a more efficient and generalized version of FedSGD by granting
the clients to train their local model parameters on multiple batches of the local data before sharing the parameters of the model with the global server, instead of sharing the gradients as in FedSGD. 
Hence, FedAvg can be defined as the weighted average of the local updated models. 

\textbf{Model poising attacks:}
In distributed learning setup, the global model is aggregated from local models trained by honest participants. 
However, in a real world scenario, data may contain biases, such as wrongly annotated or missing labels or missing data. 
Such faulty clients sabotage the learning process by sending bad local models to the global server.
In this work, we highlight two main strategies for bad clients, and assess their impact on the performance of the FL pipeline, namely:
\begin{itemize}
    \item \emph{Faulty or Byzantine}: Clients that add random noise to the model parameters before sending them back to the aggregator.
    \item \emph{Malicious}: Clients that train their models in the same way as healthy clients, but instead of using correct labels they purposely train on incorrect labels, known as a label flipping attack.
\end{itemize}
Standard aggregation methods such as FedAvg are vulnerable to these bad local models.

\textbf{Robust aggregation:}
Several works have been published to tackle the aforementioned drawbacks of FedAvg against adversaries. 
These methods rely mainly on the distribution of the local models or the distance between local and global models.
For example, COMED \cite{yin2018byzantine} utilises the coordinate-wise median of the client model parameters to construct the global model.
Other methods, such as Krum, Multi-Krum \cite{blanchard2017machine}, AFA \cite{munoz2019byzantine} and FedMGDA+ \cite{hu2020fedmgda+} assigns a score to each client depending on the similarity of the models. 
These often consist of using a metric, such as cosine similarity or L2 distance between local and global model parameters. 
These scores are then used to determine which models are averaged and in what proportions.
AFA and FedMGDA+ can also block clients permanently if they consistently are assigned a low score every federated round.

\textbf{Data heterogeneity:}
In real-world scenario, participants of a distributed learning process are independent and have different distribution characteristics. 
The majority of robust aggregation methods methods are optimized to work on independently and identically distributed (IID) data between the clients.
Although such methods can detect and block bad participants from the model aggregation step, benign clients with different data or labels than the majority will be also blocked and excluded during model aggregation.
Inspired by \cite{kairouz2019advancesOpenProblems}, we adapt \emph{label skewed} and \emph{quantity skewed} data for the non-IID participants, simulated from IID datasets using the Dirichlet distribution. 


\textbf{Knowledge distillation:}
In this work, we utilize knowledge distillation for robust model aggregation. 
Hinton et al. \cite{hinton2015distilling} showed that soft predictions from a teacher model, or an ensemble of teacher models, can be used to train a student model.
Recently, Lin et al. \cite{lin2020ensembleFedDF} proposed a method, FedDF, that uses Knowledge distillation in federated setting to learn a global model from an ensemble of client models using model fusion. 
A similar work by Chen et al. \cite{chen2020fedbe}, called FedBE, also uses knowledge distillation to teach a student model to match predictions from a Bayesian model ensemble.
One drawback of FedDF and FedBE is that the aggregation server needs to have its own unlabeled dataset. However, this can be overcome using zero-shot learning, where synthetic data can be generated for this purpose \cite{micaelli2019zeroshot, nayak2019zeroshot2}.

\section{Federated robust adaptive distillation (FedRAD)}
In this paper, knowledge distillation is performed using an ensemble of models to construct soft pseudolabels for an unlabeled dataset on the server side. 
A set of logits is predicted using the unlabeled data as an input of each model, from which the pseudolabels are constructed. 
FedBE applies the softmax function first to get the pseudo-probabilities from each model and then average those to make the pseudolabels,
while FedDF instead averages the logits before applying a softmax function, resulting in soft pseudo-probabilities that can be used as labels during training. 
This works well when there are no attacks. However, when one or more classifiers are confidently predicting the incorrect label, it can affect the average of the logits, as shown in Table \ref{tab:medianRobust}.
Since logits outputs are unbounded, a single attacker can shift the average significantly. 
To tackle this drawback, we make use of the robust statistical characteristics of the median instead of the mean. 
This improves the robustness of the pseudolabels against outliers from the predicted models' logits.

\begin{figure}[ht]
    \centering
    \adjustbox{max width=\textwidth}{
        \includegraphics[angle=270, trim={100 100 100 100}, clip=true]{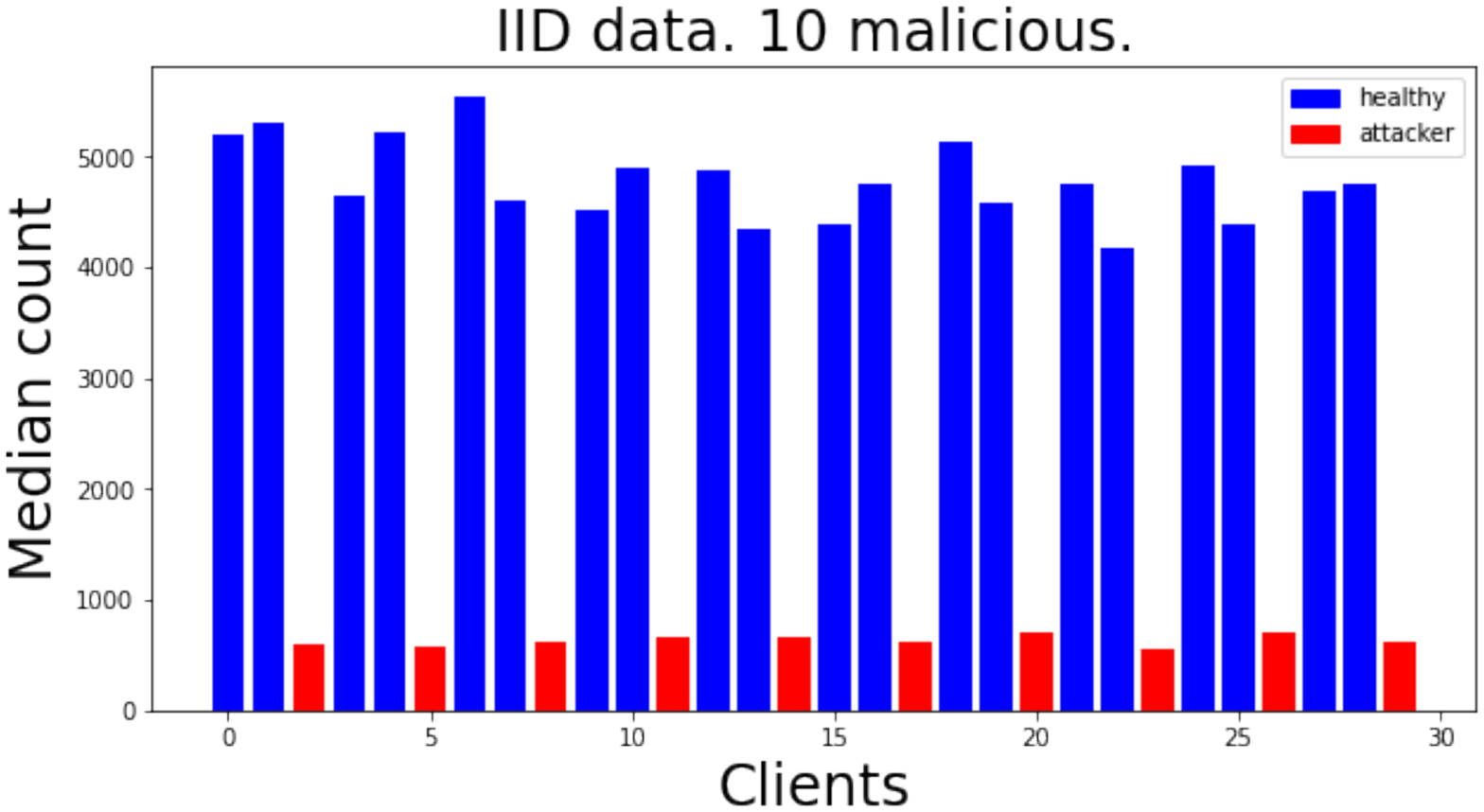}
        \includegraphics[angle=270, trim={100 100 100 100}, clip=true]{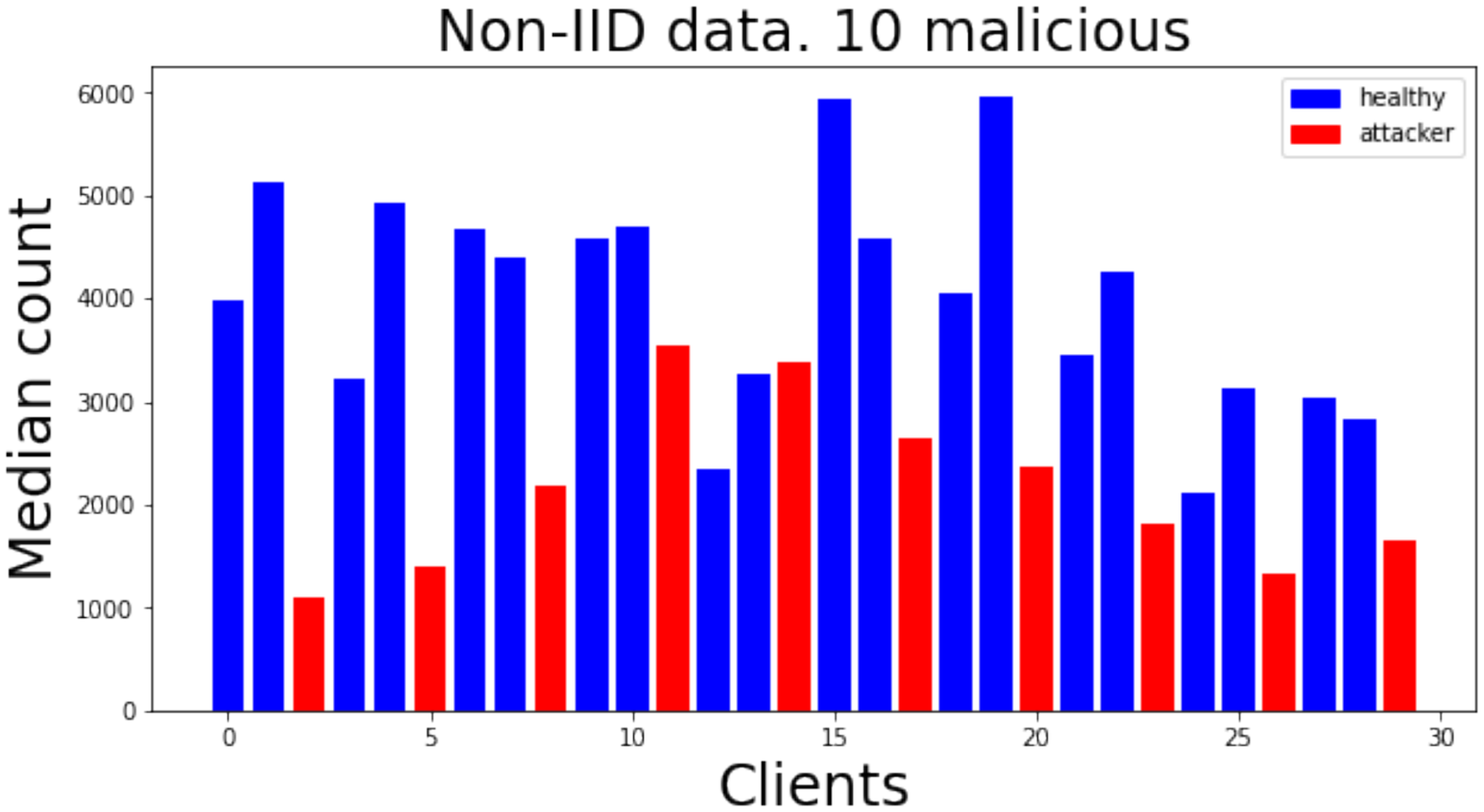}
        \includegraphics[angle=270, trim={100 100 100 100}, clip=true]{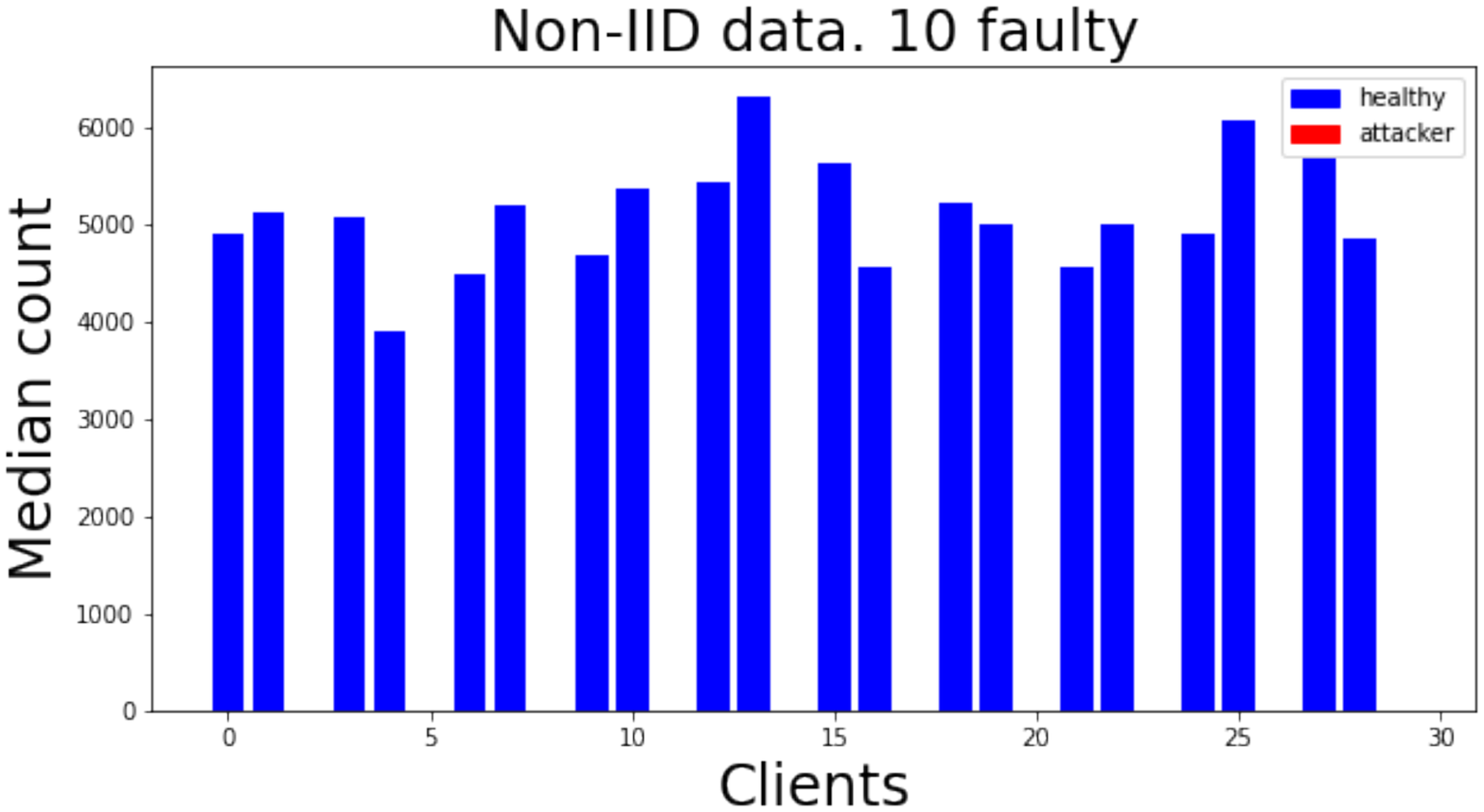}
    }
    \caption[Median-counting Histograms]{Median-counting histograms for prediction on $10,000$ images from the MNIST dataset. There are 30 clients, 10 of which are attacking. Attackers are clients number $2, 5, 8, 11, 14, 17, 20, 23, 26, 29$. Left: IID, 10 Malicious. Middle: Non-IID, 10 Malicious. Right: Non-IID, 10 Faulty.}
    \label{fig:median-histograms}
\end{figure}

The median logits pseudolabels are analysed empirically against attackers using the server-side unlabelled dataset. This can be done by counting how often each client's model has a logit output that is selected as the median value. 
Histograms of how often each client's model had the median output are shown in Figure \ref{fig:median-histograms}. This is a 10-class prediction task on $10,000$ images from the server-side dataset, so there total count is $10\cdot 10,000 = 100,000$.
The Faulty clients, i.e. ones that send back noisy updates, have a count close to $0$. The malicious clients are on average chosen less often than the healthy clients, even in non-IID settings. This suggests that the pseudo-labels which are constructed using median logits will be less affected by faulty and malicious clients. By training the averaged model to match these pseudolabels, the effects of adversarial models is dampened. 

\textbf{Scoring system using medians:}
The median-based scoring algorithm assigns each client a score based on the frequency of which it outputs the median logit for a class prediction. This is done by counting how many times each client gave the median logit prediciton, and then normalizing the counts such that the scores add up to $1$, see Algorithm \ref{alg:MedianScores}. 
These scores are then used for the weighted average of the client models. 
Since the median-based scores are on average lower for bad clients than healthy ones, this should decrease the effect of the bad model updates. 


\begin{algorithm}
\caption{Median-based scores (MedianScores)}\label{alg:MedianScores}
\begin{algorithmic}
\Require An ensemble $E = \{m_i\}_{i=1}^M$ of $M$ models for $C$-dimensional classification problem. A dataset $D$ (unlabeled).
\Ensure MedianScores$(E, D)$:
\State $P = \{0\}_{i=1}^M$  \Comment{Counter}
\For{each datapoint $d \in D$}
    \For{model $m \in E$}
        \State $\hat y_m \gets m(d)$ \Comment $(1\times C)$-dimensional logits output of model $m$ for input $d$.
    \EndFor
    \State $med, idx \gets median( \{\hat y_m\}_{m\in E}, dim=0)$ \Comment Median in ensemble dimension
    \For {$i$ in $idx$}
        \State $P_i \gets P_i + 1$ \Comment{Count how often model number $i$ was the median}
    \EndFor
\EndFor
\State $P \gets P/sum(P)$ \Comment{Normalizing the scores such that $\sum_i P_i = 1$}
\State \textbf{return} $P$ 
\end{algorithmic}
\end{algorithm}



\textbf{FedRAD:}
The aggregation server starts by initializing the weights of a global model. Then in each round, a subsample of clients are chosen and the global model is sent to them. Each client trains the global model using their data, then sends the model back to the aggregation server.
Once the aggregation server has all the models, the unlabelled dataset is used to compute the median-based scores using Algorithm \ref{alg:MedianScores}. These scores can be adjusted by the size of each client's dataset. A weighted average is then taken of the client models to construct a new student model.
Finally, the student model is trained using median-based knowledge distillation, and assigned to the new global model. These steps are repeated in federated rounds until convergence. 
Algorithm \ref{alg:FedRAD} demonstrates the pseudocode of the proposed FedRAD.

\begin{algorithm}[ht]
\caption{Federated Robust Adaptive Distillation (FedRAD)}\label{alg:FedRAD}
\begin{algorithmic}
\Require A set of participating clients $S = \{k_i\}_{i=1}^M$, each of which has a dataset of size $n_k$. Unlabelled server-side dataset $D_s$.
\Ensure FedRAD:
\State Initialise a global model with weights $\boldsymbol w_0^g$.
\For{each round $t=1,\dots,T$}
    \State $S_t \gets$ Sample a subset of $C$ fraction of clients from $S$
    \State Send global model weights $\boldsymbol w_{t-1}^g$ to clients in $S_t$
    \For{each client $k \in S_t$ \textbf{in parallel}}
        \State $\boldsymbol{w}_t^k \gets $ Updated global model using client data
    \EndFor
    \State $\boldsymbol{p}_t \gets$ MedianScores$(\{\boldsymbol{w}_t^k\}_{k\in S_t}, D_s)$ \Comment{See algorithm \ref{alg:MedianScores}}
    \State $\boldsymbol{p}_t \gets \left\{ \frac{n_k \cdot p_t^k}{\sum_k n_k \cdot p_t^k} \right\}_{k\in S_t}$ \Comment{Adjust by size of datasets}
    \State $\boldsymbol w_t^g \gets \sum_{k\in S_t} p_t^k \cdot \boldsymbol{w}_t^k$ \Comment{Element-wise weighted average}
    \State $\boldsymbol w_t^g \gets $ KnowledgeDistillation$(\boldsymbol w_t^g,  \{\boldsymbol{w}_t^k\}_{k\in S_t}, D_s)$ \Comment{Median-based KD}
\EndFor
\end{algorithmic}
\end{algorithm}

\section{Experiments and results}\label{sec:experiments}

\textbf{Simulating non-IID data:}
Here, we use the Dirichlet distribution which is a multivariate continuous probability distribution. 
It is parameterized by a K-dimensional vector $\boldsymbol{\alpha} = \{\alpha_i\}_{i=1}^K, \forall i: \alpha_i > 0$.
Its Probability Density Function (PDF) is 
$$f(\boldsymbol X ; \boldsymbol \alpha) = \frac{1}{B(\boldsymbol \alpha)} \prod_{i=1}^K x_i^{\alpha_i-1}$$
where $B(\boldsymbol \alpha)$ is a normalizing constant and $\Gamma(z) = \int_0^\infty x^{z-1} e^{-x}dx$ is the gamma function.

Samples from the Dirichlet distributions are $K$-dimensional vectors. The support is the open standard $(K-1)$-simplex, $\{x\in \mathbb{R}^K: x_i \in (0,1) \text{ for all }i = 1,2,...,K ; \sum_{i=1}^K x_i = 1\}$.

\begin{figure}[ht]
\centering
    \adjustbox{max width=\textwidth}{
        \includegraphics[width=0.33\textwidth, trim={100 20 100 20}, clip=true]{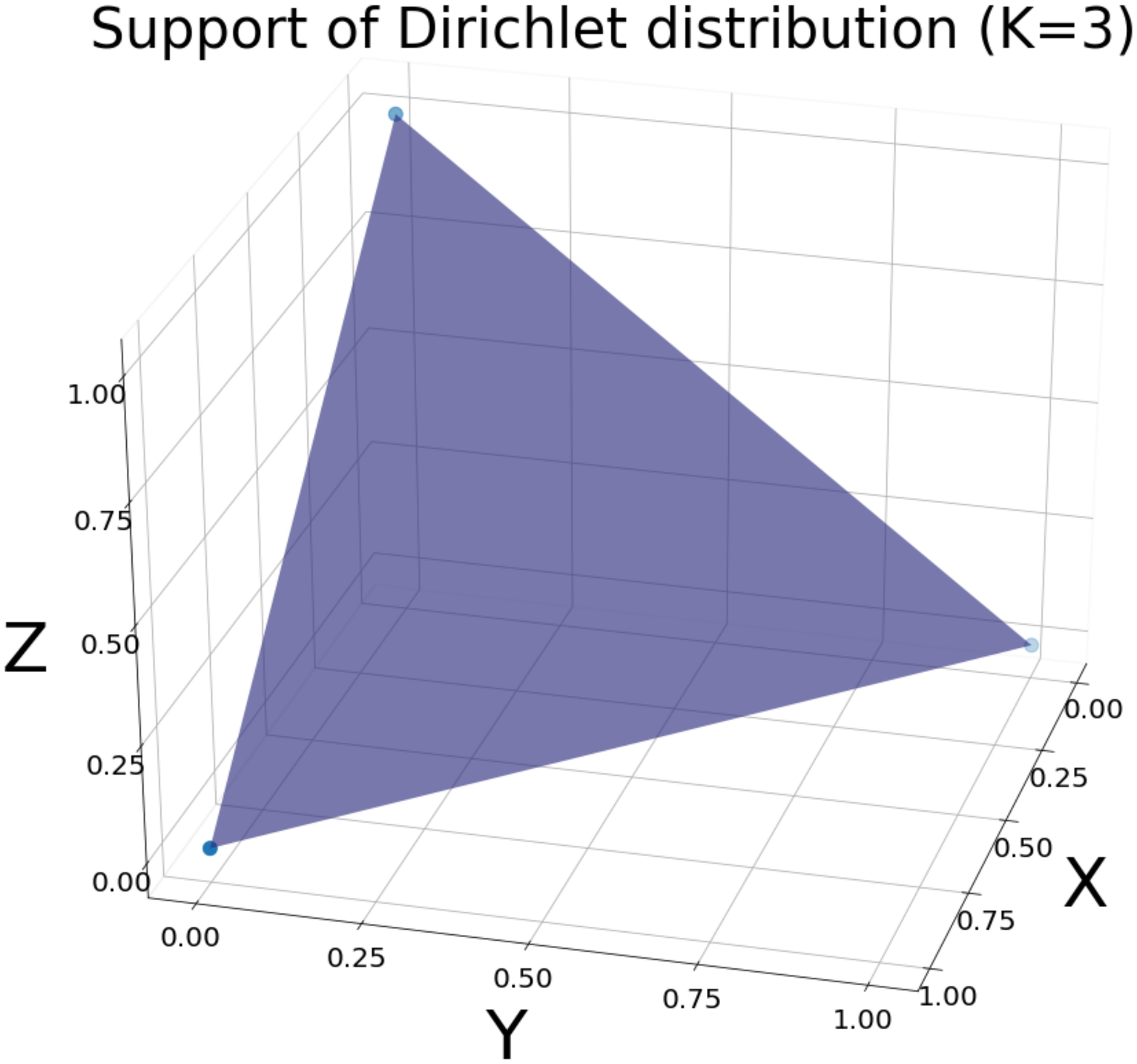}
        \includegraphics[width=0.33\textwidth, trim={100 300 100 300}, clip=true]{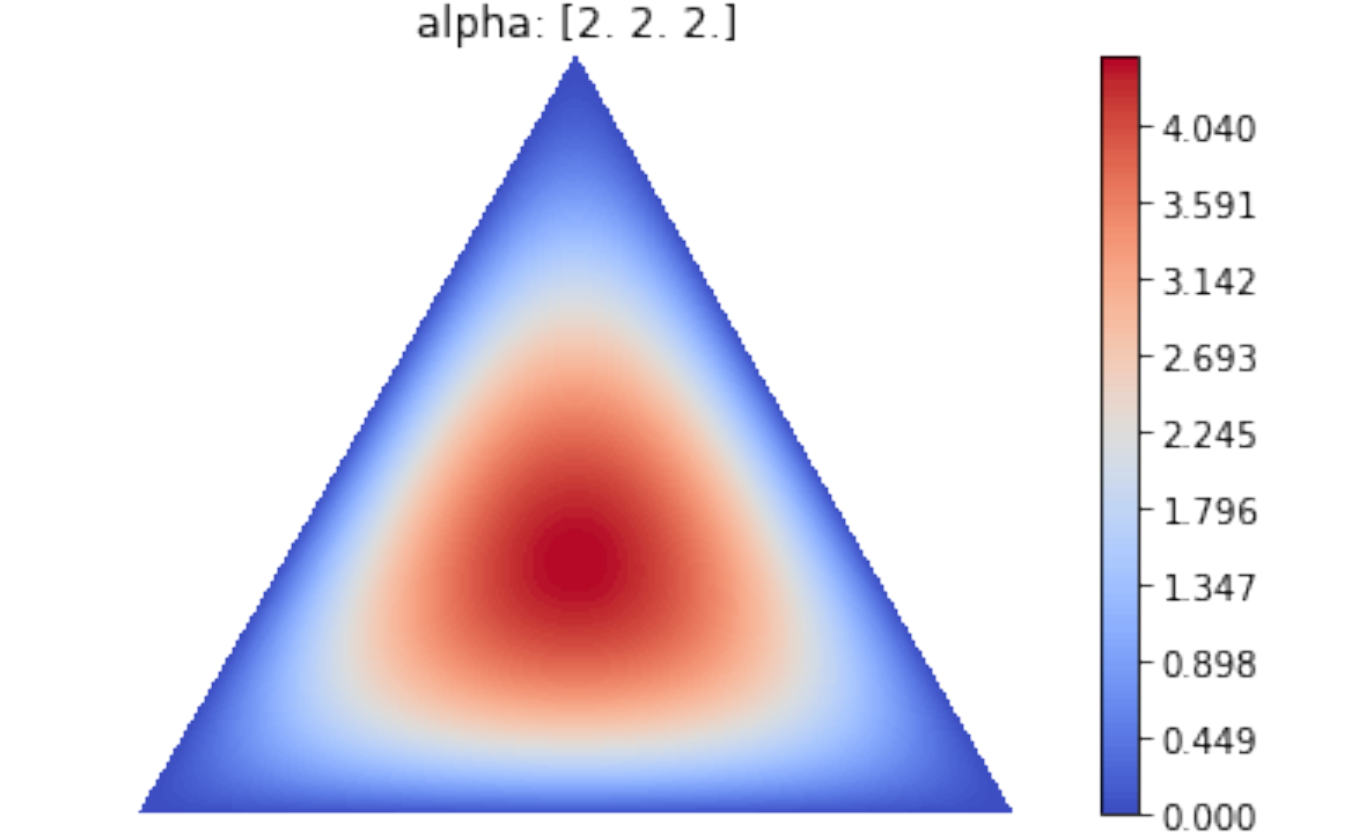}
        \includegraphics[width=0.33\textwidth, trim={100 300 100 300}, clip=true]{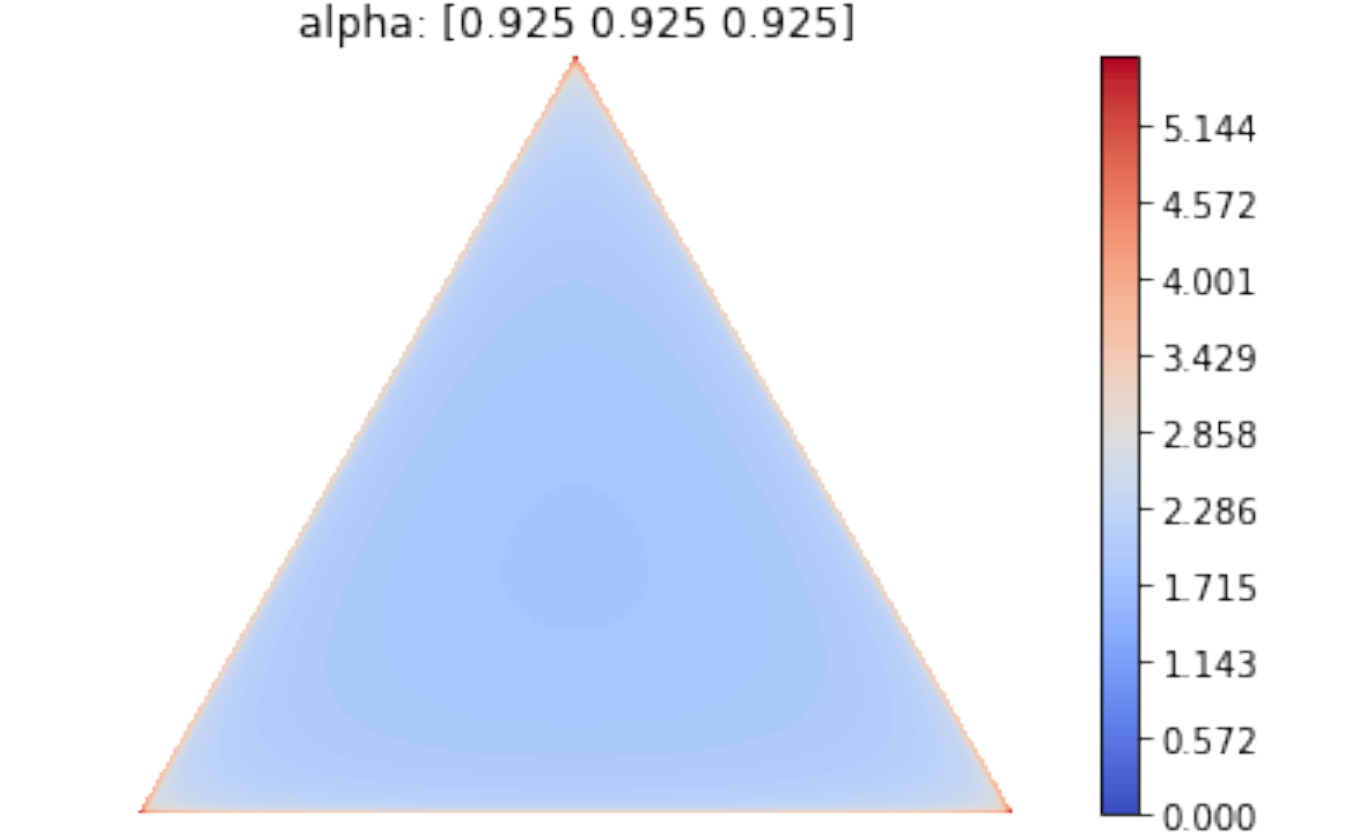}
        }
        \caption[Dirichlet Distribution - Support and PDFs]{Support and Contour maps of PDFs of Dirichlet distributions (K=3). Left: Support. Middle: Symmetrical $\alpha>1$. Right: Symmetrical $\alpha<1$. }
\label{fig:dirichlets}
\end{figure}

A symmetrical Dirichlet distribution can be used to simulate a non-IID split of datasets between agents. This is done by distributing the samples in each class unequally between agents, in proportions sampled from a symmetrical Dirichlet distribution. 
If we have a dataset with $C$ classes and want to divide the data among $K$ different agents, we can define a symmetrical $K$-dimensional Dirichlet distribution $Dir(\boldsymbol{\alpha})$ where $\boldsymbol{\alpha} = \alpha \cdot \boldsymbol{1} = \{\alpha\}_{i=1}^K$.
We then sample $C$ vectors from the distribution, one for each class, and divide the data in each class between the agents in proportion with the sampled values. 
The $\alpha$ hyperparameter can be tuned to control how unequally the data classes are divided between agents. A high value of $\alpha$ leads to a low variance in proportion, resulting in more equal splits between agents, whereas a low $\alpha$ increases the variance of the proportions, leading to a more non-IID data split. 



\textbf{Implementation:}
In order to thoroughly assess the performance of FedRAD, we have implemented several aggregation methods and two types of adversarial attacks. 
\emph{Faulty} clients are simulated by adding Gaussian noise with a variance of $20$ to the model parameters. 
\emph{Malicious} clients are simulated by setting all their data labels to $0$ before training.

\begin{figure}[ht] 
\centering
    \adjustbox{max width=\textwidth}{
        \includegraphics[trim={100 120 100 120}, clip=True]{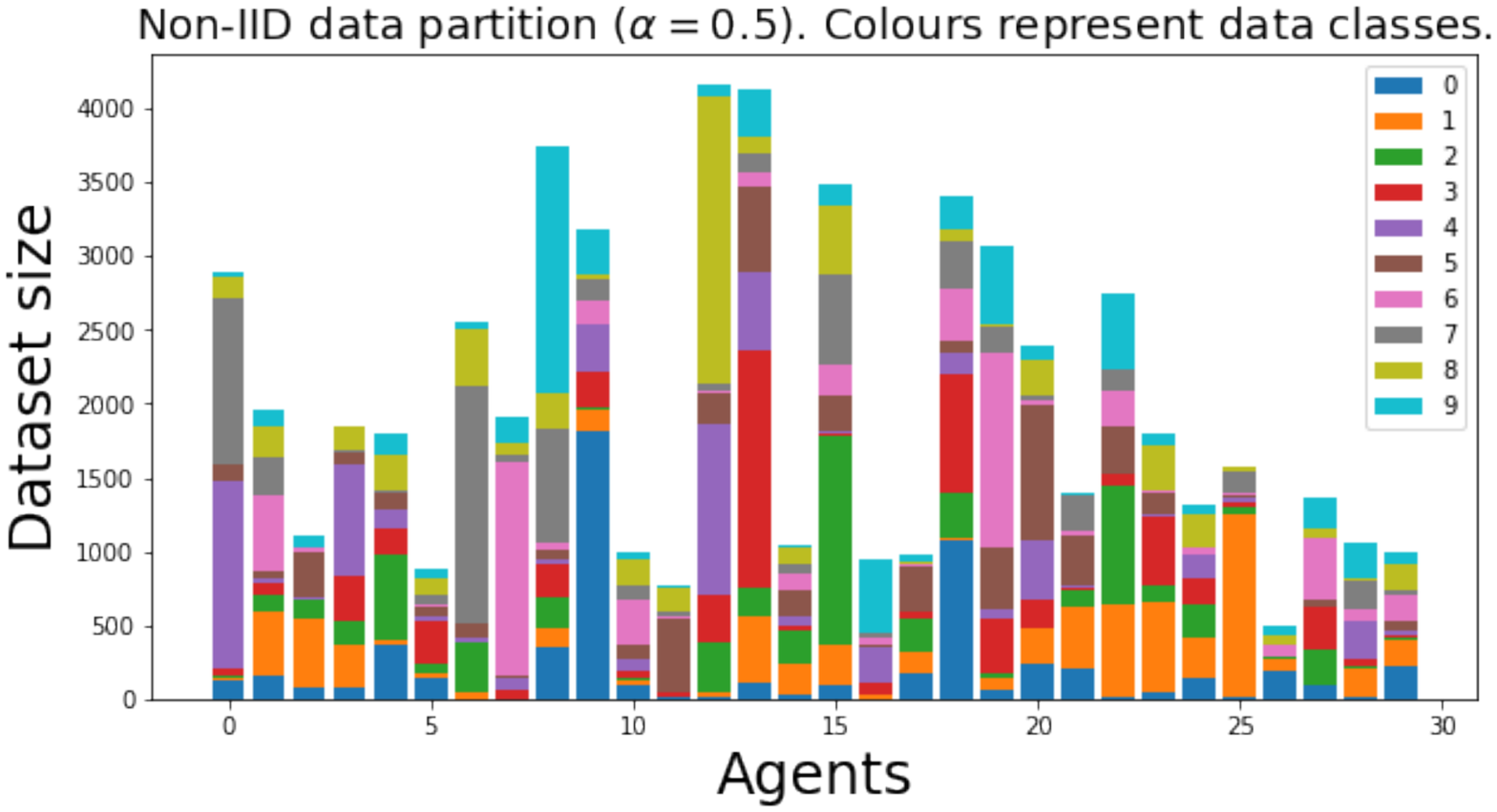}
        \includegraphics[trim={100 120 100 120}, clip=True]{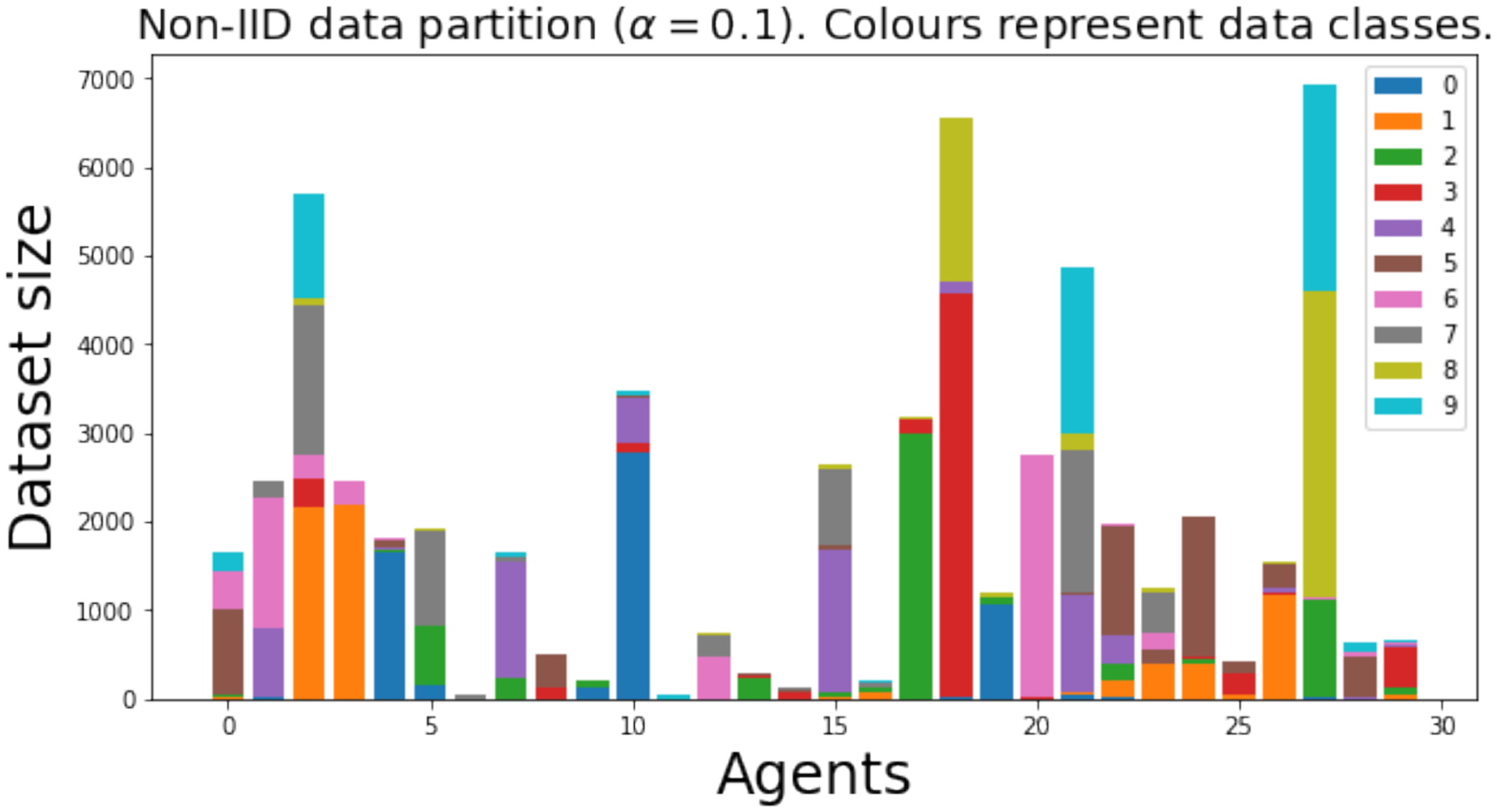}
    }
\caption[Simulated Non-IID Data]{Examples of simulated non-IID data splits using samples from a symmetrical Dirichlet distribution. Left: $\alpha=0.5$. Right: Symmetrical $\alpha=0.1$.}
\label{fig:data_nonIID}
\end{figure}

Non-IID data distributions are simulated with three levels of heterogeneity: IID data split with slight quantity skews, a slightly non-IID split sampled from a symmetrical Dirichlet distribution with $\alpha=0.5$ and a very non-IID split which uses $\alpha=0.1$, see Figure \ref{fig:data_nonIID}.

A knowledge distillation module is implemented by training a student model to emulate predictions from an ensemble of models. Two aggregation methods which use Knowledge Distillation, FedDF \cite{lin2020ensembleFedDF} and FedBE \cite{chen2020fedbe}, are also implemented.

For our experiments, we use the MNIST dataset \cite{lecun1998mnist}, which is a labeled image dataset of handwritten digits from 0 to 9. It has a total of 70 thousand images of size $28\times 28$. The training set includes 60 thousand images, and the test set has the remaining 10 thousand. 
There are roughly the same number of images in each class.
Our classifier consists of a 3-layer fully-connected neural network that outputs the logits it assigns to each of the 10 classes. A Softmax function is then used to get the predicted probabilities.
We ran our experiments for 30 communication rounds, with 5 local epochs and a local learning rate of $0.05$. 
Performance is measured as the Error rate calculated on the test set.

FedRAD code and implementation is publicly available on GitHub
\footnote{\url{https://github.com/stefansturlu/FederatedMedical}}. 


\textbf{Results:}
The novel aggregators proposed in this project, median-based FedDF (FedDFmed) and FedRAD, outperformed all other aggregators in the presence of attacks, both in IID and non-IID settings.
Using medians instead of average logits for knowledge distillation is effective to combat malicious label-flipping attacks. 
Furthermore, the median-based scoring system enhances the robustness of the trained models by detecting faulty attackers and assigning them very low scores. 
It is also successful in giving lower scores to malicious attackers than healthy agents. FedRAD combines the novel median-based knowledge distillation with the novel median-based scoring system.

Table \ref{tab:final_results} shows that it is more difficult to detect malicious attacks than faulty ones by examining the model parameters, as malicious agents use the same training procedure as healthy clients and aren't extremely noisy like the faulty models. 
The model weights of malicious agents are similarly distributed to healthy ones, and in non-IID settings it becomes difficult to detect malicious agents using similarity metrics.
Table \ref{tab:final_results} also demonstrates that FedRAD under a variety of attacks and data heterogeneity outperforms all the other aggregation methods. 




\begin{table}[ht]
    \centering
    \begin{tabular}{|p{2.0cm}| p{2.2cm}|>{\raggedleft\arraybackslash}p{2.4cm}|>{\raggedleft\arraybackslash}p{2.4cm}|>{\raggedleft\arraybackslash}p{2.4cm}|}
    \hline
    \multirow{3}{*}{Attacks}  & \multirow{3}{*}{Aggregator} & \multicolumn{3}{c|}{Error rates ($\%$)} \\ \cline{3-5}
& & \multicolumn{1}{c}{\multirow{2}{*}{IID}} & \multicolumn{2}{|c|}{non-IID} \\ \cline{4-5}
& & & \multicolumn{1}{c|}{$\alpha=0.5$} & \multicolumn{1}{c|}{$\alpha=0.1$} \\ \hline \hline
    
    \multirow{8}{2.0cm}{No attacks}
     & FA & $\boldsymbol{5.24\pm0.07}$ & $\boldsymbol{5.68\pm0.11}$ & $\boldsymbol{9.13\pm0.82}$ \\ \cline{2-5}
     & COMED & $\boldsymbol{5.30\pm0.08}$ & $6.38\pm0.18$ & $81.15\pm12.94$ \\ \cline{2-5}
     & MKRUM & $5.68\pm0.16$ & $7.63\pm0.62$ & $40.93\pm13.51$ \\ \cline{2-5}
     & AFA & $\boldsymbol{5.28\pm0.17}$ & $5.94\pm0.16$ & $\boldsymbol{10.36\pm0.96}$ \\ \cline{2-5}
     & FedMGDA++ & $6.28\pm0.25$ & $12.15\pm5.39$ & $21.12\pm3.84$ \\ \cline{2-5}
     & FedDF & $5.43\pm0.10$ & $\boldsymbol{5.82\pm0.15}$ & $\boldsymbol{9.17\pm0.40}$ \\ \cline{2-5}
     & FedDFmed & $5.45\pm0.08$ & $\boldsymbol{5.86\pm0.25}$ & $\boldsymbol{9.24\pm0.43}$ \\ \cline{2-5}
     & FedRAD & $\boldsymbol{5.29\pm0.16}$ & $\boldsymbol{5.82\pm0.15}$ & $\boldsymbol{9.38\pm0.57}$ \\ 
     \hline \hline
    
    \multirow{8}{2.0cm}{10 Faulty}
     & FA & $90.03\pm0.96$ & $89.58\pm0.50$ & $90.10\pm0.63$ \\ \cline{2-5}
     & COMED & $\boldsymbol{5.39\pm0.14}$ & $6.44\pm0.28$ & $69.22\pm22.75$ \\ \cline{2-5}
     & MKRUM & $89.66\pm0.63$ & $90.37\pm0.37$ & $90.03\pm0.65$ \\ \cline{2-5}
     & AFA & $70.97\pm32.56$ & $90.12\pm0.47$ & $89.70\pm0.33$ \\ \cline{2-5}
     & FedMGDA++ & $89.82\pm0.71$ & $89.37\pm0.28$ & $89.89\pm0.43$ \\ \cline{2-5}
     & FedDF & $90.06\pm0.82$ & $89.65\pm1.01$ & $89.71\pm0.62$ \\ \cline{2-5}
     & FedDFmed & $81.02\pm3.56$ & $90.14\pm0.54$ & $90.11\pm0.61$ \\ \cline{2-5}
     & FedRAD & $\boldsymbol{5.44\pm0.07}$ & $\boldsymbol{5.91\pm0.31}$ & $\boldsymbol{12.63\pm3.65}$ \\ 
     \hline \hline
    
    \multirow{8}{2.0cm}{10 Malicious}
     & FA & $17.33\pm3.64$ & $74.87\pm30.66$ & $38.09\pm27.79$ \\ \cline{2-5}
     & COMED & $25.11\pm1.39$ & $48.51\pm21.52$ & $90.08\pm0.40$ \\ \cline{2-5}
     & MKRUM & $31.15\pm25.15$ & $10.82\pm2.19$ & $60.78\pm15.24$ \\ \cline{2-5}
     & AFA & $17.68\pm4.46$ & $90.20\pm0.00$ & $90.20\pm0.00$ \\ \cline{2-5}
     & FedMGDA++ & $90.20\pm0.00$ & $90.20\pm0.00$ & $90.20\pm0.00$ \\ \cline{2-5}
     & FedDF & $65.20\pm28.69$ & $76.12\pm28.16$ & $61.49\pm31.78$ \\ \cline{2-5}
     & FedDFmed & $8.34\pm0.25$ & $9.24\pm1.23$ & $16.74\pm3.87$ \\ \cline{2-5}
     & FedRAD & $\boldsymbol{5.89\pm0.20}$ & $\boldsymbol{6.55\pm0.43}$ & $\boldsymbol{13.09\pm1.97}$ \\ 
     \hline \hline
    
    \multirow{8}{2.0cm}{5 Faulty, 5 Malicious}
     & FA & $89.64\pm0.55$ & $89.75\pm0.75$ & $89.62\pm0.63$ \\ \cline{2-5}
     & COMED & $8.00\pm0.42$ & $10.21\pm0.93$ & $89.10\pm0.90$ \\ \cline{2-5}
     & MKRUM & $89.60\pm0.36$ & $89.66\pm0.43$ & $89.82\pm0.67$ \\ \cline{2-5}
     & AFA & $67.62\pm6.64$ & $67.82\pm30.59$ & $80.97\pm17.03$ \\ \cline{2-5}
     & FedMGDA++ & $90.20\pm0.00$ & $90.21\pm0.00$ & $89.31\pm1.61$ \\ \cline{2-5}
     & FedDF & $90.39\pm0.57$ & $90.17\pm0.67$ & $89.97\pm0.18$ \\ \cline{2-5}
     & FedDFmed & $55.11\pm5.41$ & $90.20\pm0.00$ & $90.20\pm0.00$ \\ \cline{2-5}
     & FedRAD & $\boldsymbol{5.63\pm0.20}$ & $\boldsymbol{6.15\pm0.40}$ & $\boldsymbol{10.95\pm1.22}$ \\ \cline{2-5}
     \hline
    \end{tabular}
    
    \vspace{0.1cm}
    \caption[Summary of Results for all Aggregation Methods]{Test set error rate for MNIST after 30 rounds. Average and standard deviation of test-set error rates shown for $5$ different random seeds. The best performing methods in each category, and those not statistically significantly different according to a t-test, are in bold.}
\label{tab:final_results}
\end{table}

\begin{figure}[ht] 
\centering
    \adjustbox{max width=\textwidth}{
        \includegraphics[trim={95 250 95 250}, clip=True]{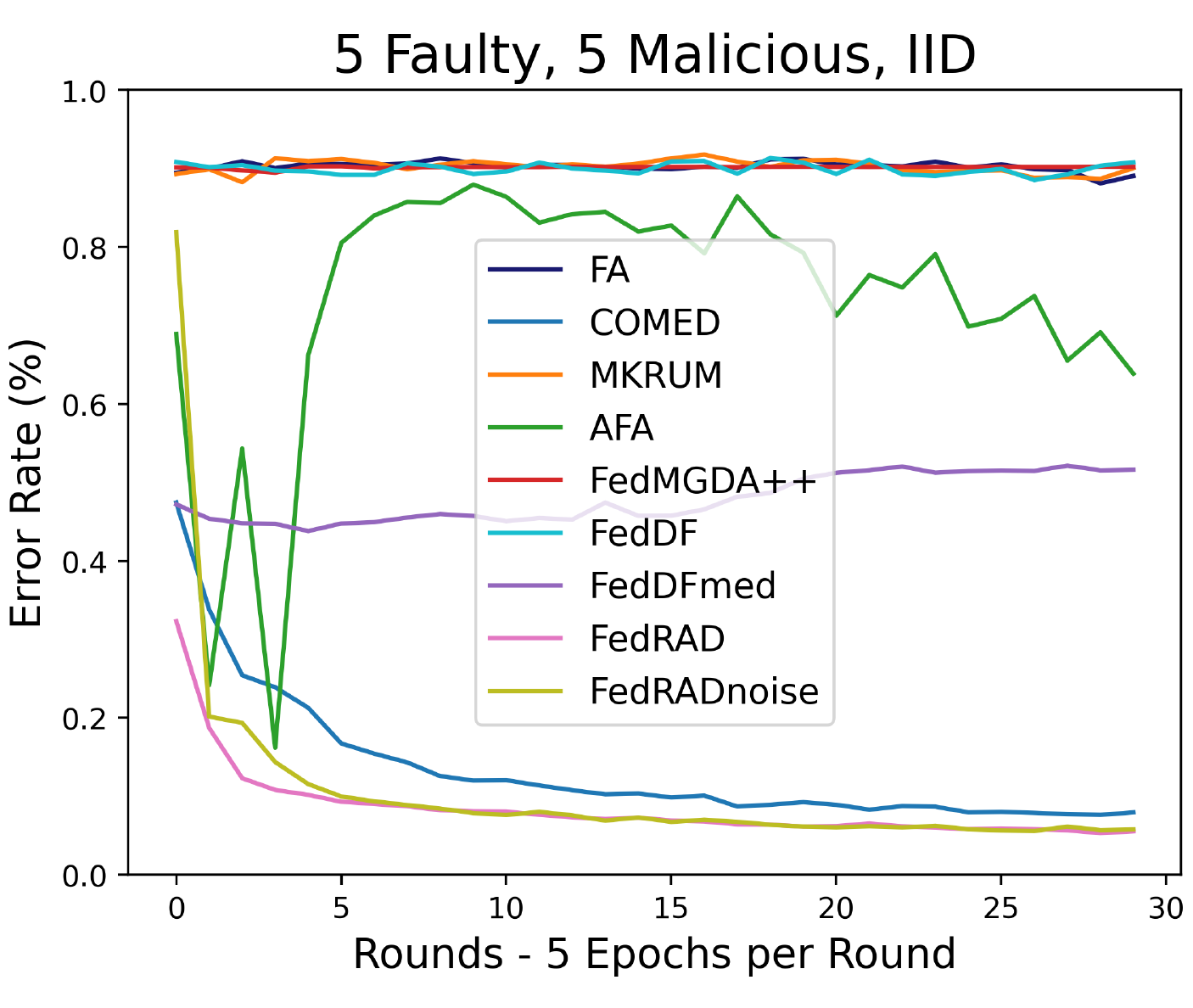}
        \includegraphics[trim={95 250 95 250}, clip=True]{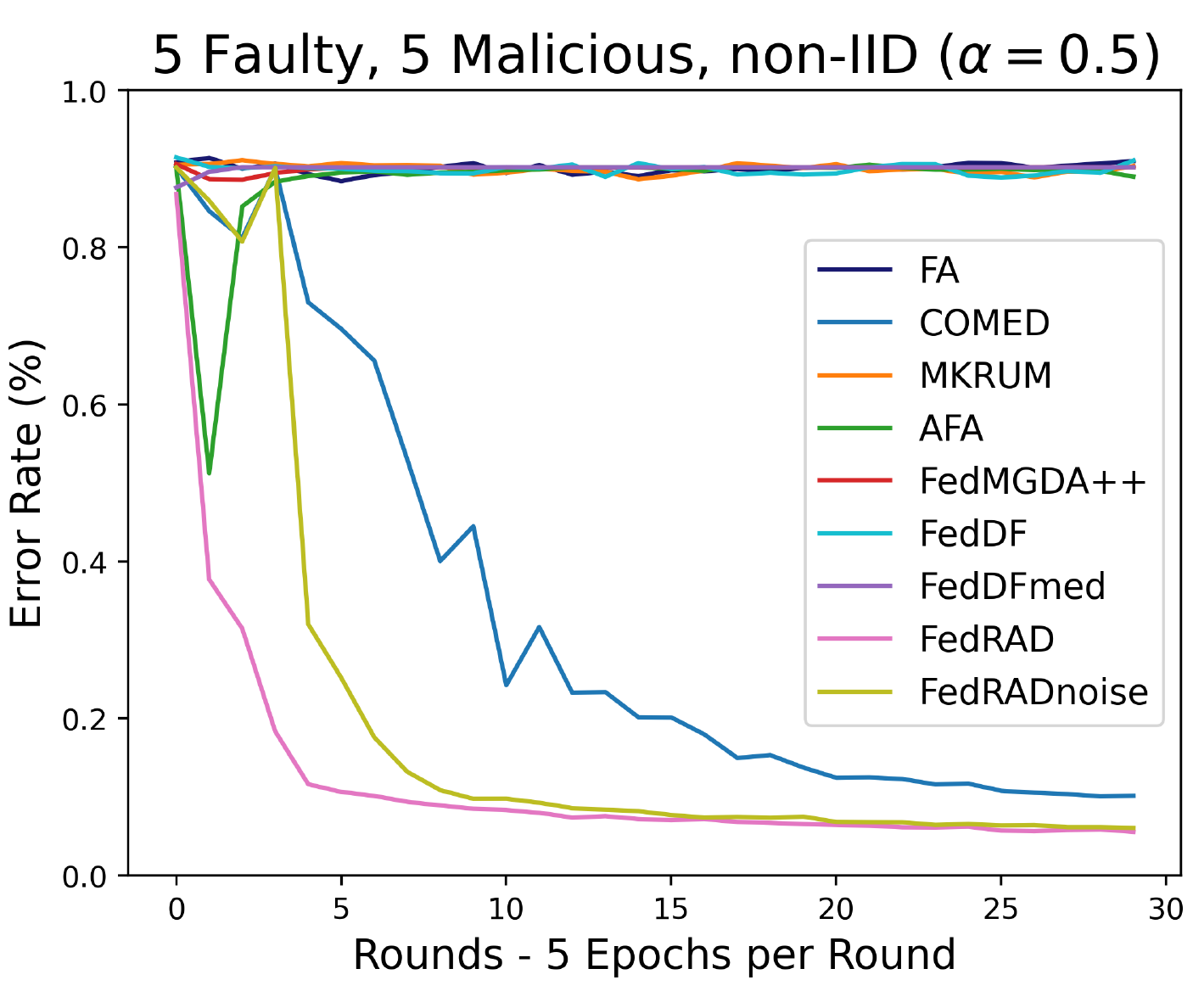}
        \includegraphics[trim={95 250 95 250}, clip=True]{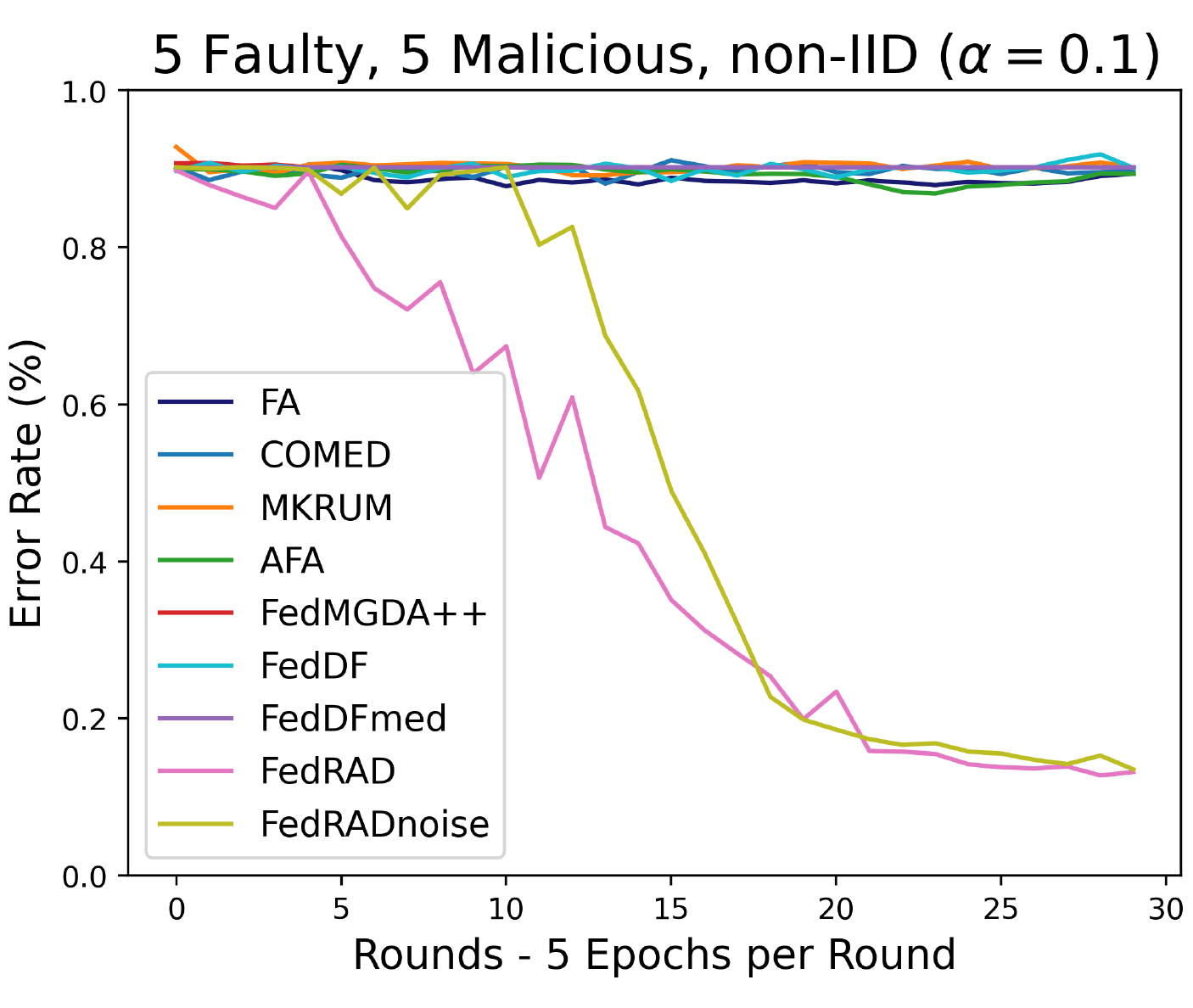}
    }
\caption{Effects of non-IID data with $5$ faulty and $5$ malicious attackers.}
\label{fig:attack_non-iid_dual5}
\end{figure}

Figure \ref{fig:attack_non-iid_dual5} displays the learning curves in experiments with both faulty and malicious attackers in IID and non-IID settings. 
With 5 faulty and 5 malicious attacks, COMED did well in IID scenarios, but as usual failed for non-IID scenarios. However, FedRAD consistently performed well under both types of attacks at the same time.
More results can be found in Appendix \ref{app:learningCurves}.

\section{Conclusions}
In this work, we propose FedRAD for robust federated aggregation using a scoring system based on the median statistic and network distillation.
In contrary to previous robust aggregation methods that use the high-dimensional vectors of parameters of the models, our method use the output predictions from the models for detecting adversaries in the presence of heterogeneous client data.

Extensive experiments were conducted using the MNIST dataset to evaluate the performance of different baseline methods in the presence of attackers, for both IID and non-IID client data.
Existing aggregators, like COMED and Multi-Krum, were shown to break down when client data is heterogeneous. Existing adaptive aggregators, AFA and FedMGDA+, were also shown to be unable to detect and block adversaries in non-IID settings.
 
To conclude, FedRAD outperforms all other aggregation methods in the presence of attacks, and is the only aggregator to successfully combat adversaries in both IID and non-IID settings. We will investigate the performance of FedRAD with privacy preserving techniques such as differential privacy in the future.

\begin{ack}
\textbf{Acknowledgments}
This work is supported by the UK Research and Innovation London Medical Imaging and Artificial Intelligence Centre for Value Based Healthcare, and in part by the Intramural Research Program of the National Institutes of Health Clinical Center.

\end{ack}


\bibliographystyle{unsrtnat}
\bibliography{NFFL2021}

\begin{thebibliography}{20}
\providecommand{\natexlab}[1]{#1}
\providecommand{\url}[1]{\texttt{#1}}
\expandafter\ifx\csname urlstyle\endcsname\relax
  \providecommand{\doi}[1]{doi: #1}\else
  \providecommand{\doi}{doi: \begingroup \urlstyle{rm}\Url}\fi

\bibitem[Krizhevsky et~al.(2012)Krizhevsky, Sutskever, and
  Hinton]{krizhevsky2012imagenet}
Alex Krizhevsky, Ilya Sutskever, and Geoffrey~E Hinton.
\newblock Imagenet classification with deep convolutional neural networks.
\newblock \emph{Advances in neural information processing systems},
  25:\penalty0 1097--1105, 2012.

\bibitem[Jiang et~al.(2017)Jiang, Jiang, Zhi, Dong, Li, Ma, Wang, Dong, Shen,
  and Wang]{jiang2017healthcare}
Fei Jiang, Yong Jiang, Hui Zhi, Yi~Dong, Hao Li, Sufeng Ma, Yilong Wang, Qiang
  Dong, Haipeng Shen, and Yongjun Wang.
\newblock Artificial intelligence in healthcare: past, present and future.
\newblock \emph{Stroke and vascular neurology}, 2\penalty0 (4), 2017.

\bibitem[Bahdanau et~al.(2014)Bahdanau, Cho, and
  Bengio]{bahdanau2014machineTranslation}
Dzmitry Bahdanau, Kyunghyun Cho, and Yoshua Bengio.
\newblock Neural machine translation by jointly learning to align and
  translate.
\newblock \emph{arXiv preprint arXiv:1409.0473}, 2014.

\bibitem[Bewley et~al.(2019)Bewley, Rigley, Liu, Hawke, Shen, Lam, and
  Kendall]{bewley2019driving}
Alex Bewley, Jessica Rigley, Yuxuan Liu, Jeffrey Hawke, Richard Shen, Vinh-Dieu
  Lam, and Alex Kendall.
\newblock Learning to drive from simulation without real world labels.
\newblock In \emph{2019 International conference on robotics and automation
  (ICRA)}, pages 4818--4824. IEEE, 2019.

\bibitem[Obermeyer and Emanuel(2016)]{obermeyer2016dataHungry}
Ziad Obermeyer and Ezekiel~J Emanuel.
\newblock Predicting the future—big data, machine learning, and clinical
  medicine.
\newblock \emph{The New England journal of medicine}, 375\penalty0
  (13):\penalty0 1216, 2016.

\bibitem[McMahan et~al.(2017)McMahan, Moore, Ramage, Hampson, and
  y~Arcas]{mcmahan2017communication}
Brendan McMahan, Eider Moore, Daniel Ramage, Seth Hampson, and Blaise~Aguera
  y~Arcas.
\newblock Communication-efficient learning of deep networks from decentralized
  data.
\newblock In \emph{Artificial Intelligence and Statistics}, pages 1273--1282.
  PMLR, 2017.

\bibitem[Yin et~al.(2018)Yin, Chen, Kannan, and Bartlett]{yin2018byzantine}
Dong Yin, Yudong Chen, Ramchandran Kannan, and Peter Bartlett.
\newblock Byzantine-robust distributed learning: Towards optimal statistical
  rates.
\newblock In \emph{International Conference on Machine Learning}, pages
  5650--5659. PMLR, 2018.

\bibitem[Grama et~al.(2020)Grama, Musat, Mu{\~n}oz-Gonz{\'a}lez,
  Passerat-Palmbach, Rueckert, and Alansary]{grama2020robust}
Matei Grama, Maria Musat, Luis Mu{\~n}oz-Gonz{\'a}lez, Jonathan
  Passerat-Palmbach, Daniel Rueckert, and Amir Alansary.
\newblock Robust aggregation for adaptive privacy preserving federated learning
  in healthcare.
\newblock \emph{arXiv preprint arXiv:2009.08294}, 2020.

\bibitem[Blanchard et~al.(2017)Blanchard, El~Mhamdi, Guerraoui, and
  Stainer]{blanchard2017machine}
Peva Blanchard, El~Mahdi El~Mhamdi, Rachid Guerraoui, and Julien Stainer.
\newblock Machine learning with adversaries: Byzantine tolerant gradient
  descent.
\newblock In \emph{Proceedings of the 31st International Conference on Neural
  Information Processing Systems}, pages 118--128, 2017.

\bibitem[Mu{\~n}oz-Gonz{\'a}lez et~al.(2019)Mu{\~n}oz-Gonz{\'a}lez, Co, and
  Lupu]{munoz2019byzantine}
Luis Mu{\~n}oz-Gonz{\'a}lez, Kenneth~T Co, and Emil~C Lupu.
\newblock Byzantine-robust federated machine learning through adaptive model
  averaging.
\newblock \emph{arXiv preprint arXiv:1909.05125}, 2019.

\bibitem[Hu et~al.(2020)Hu, Shaloudegi, Zhang, and Yu]{hu2020fedmgda+}
Zeou Hu, Kiarash Shaloudegi, Guojun Zhang, and Yaoliang Yu.
\newblock Fedmgda+: Federated learning meets multi-objective optimization.
\newblock \emph{arXiv preprint arXiv:2006.11489}, 2020.

\bibitem[Zhao et~al.(2018)Zhao, Li, Lai, Suda, Civin, and
  Chandra]{zhao2018federatedNonIID}
Yue Zhao, Meng Li, Liangzhen Lai, Naveen Suda, Damon Civin, and Vikas Chandra.
\newblock Federated learning with non-iid data.
\newblock \emph{arXiv preprint arXiv:1806.00582}, 2018.

\bibitem[Shokri et~al.(2017)Shokri, Stronati, Song, and
  Shmatikov]{shokri2017membership}
Reza Shokri, Marco Stronati, Congzheng Song, and Vitaly Shmatikov.
\newblock Membership inference attacks against machine learning models.
\newblock In \emph{2017 IEEE Symposium on Security and Privacy (SP)}, pages
  3--18. IEEE, 2017.

\bibitem[Kairouz et~al.(2019)Kairouz, McMahan, Avent, Bellet, Bennis, Bhagoji,
  Bonawitz, Charles, et~al.]{kairouz2019advancesOpenProblems}
Peter Kairouz, H~Brendan McMahan, Brendan Avent, Aur{\'e}lien Bellet, Mehdi
  Bennis, Arjun~Nitin Bhagoji, Kallista Bonawitz, Zachary Charles, et~al.
\newblock Advances and open problems in federated learning.
\newblock \emph{arXiv preprint arXiv:1912.04977}, 2019.

\bibitem[Hinton et~al.(2015)Hinton, Vinyals, and Dean]{hinton2015distilling}
Geoffrey Hinton, Oriol Vinyals, and Jeff Dean.
\newblock Distilling the knowledge in a neural network.
\newblock \emph{arXiv preprint arXiv:1503.02531}, 2015.

\bibitem[Lin et~al.(2020)Lin, Kong, Stich, and Jaggi]{lin2020ensembleFedDF}
Tao Lin, Lingjing Kong, Sebastian~U Stich, and Martin Jaggi.
\newblock Ensemble distillation for robust model fusion in federated learning.
\newblock \emph{arXiv preprint arXiv:2006.07242}, 2020.

\bibitem[Chen and Chao(2020)]{chen2020fedbe}
Hong-You Chen and Wei-Lun Chao.
\newblock Fedbe: Making bayesian model ensemble applicable to federated
  learning.
\newblock \emph{arXiv preprint arXiv:2009.01974}, 2, 2020.

\bibitem[Micaelli and Storkey(2019)]{micaelli2019zeroshot}
Paul Micaelli and Amos Storkey.
\newblock Zero-shot knowledge transfer via adversarial belief matching.
\newblock \emph{arXiv preprint arXiv:1905.09768}, 2019.

\bibitem[Nayak et~al.(2019)Nayak, Mopuri, Shaj, Radhakrishnan, and
  Chakraborty]{nayak2019zeroshot2}
Gaurav~Kumar Nayak, Konda~Reddy Mopuri, Vaisakh Shaj, Venkatesh~Babu
  Radhakrishnan, and Anirban Chakraborty.
\newblock Zero-shot knowledge distillation in deep networks.
\newblock In \emph{International Conference on Machine Learning}, pages
  4743--4751. PMLR, 2019.

\bibitem[LeCun(1998)]{lecun1998mnist}
Yann LeCun.
\newblock The mnist database of handwritten digits.
\newblock \emph{http://yann. lecun. com/exdb/mnist/}, 1998.

\end{thebibliography}

\clearpage



\begin{appendices}

\section{Federated learning pipeline}
\begin{figure}[ht]
    \centering
    \includegraphics[width=0.6\hsize]{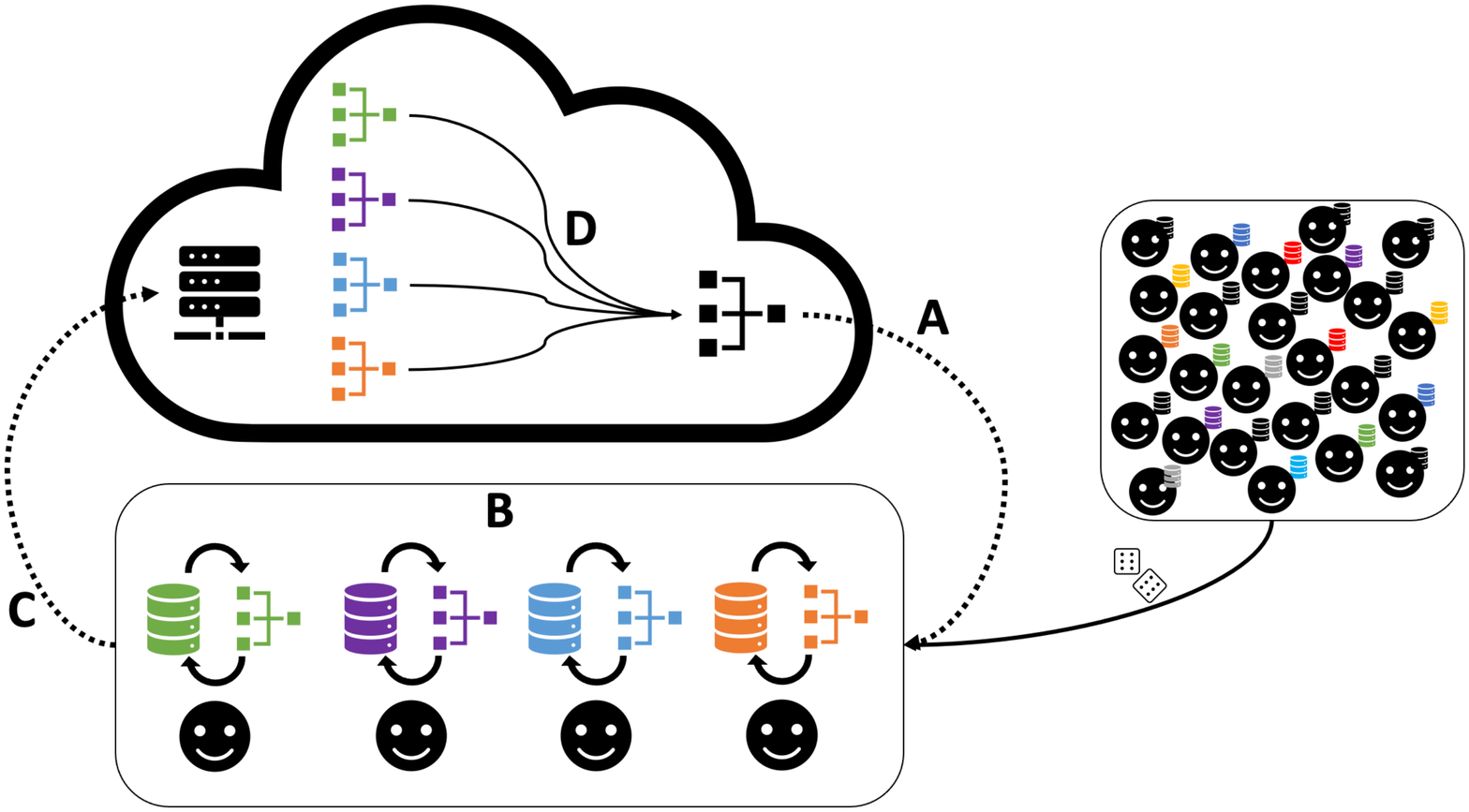}
    \caption[Federated Learning Pipeline]{Federated Learning pipeline. \textbf{A:} The global model is sent to a random subset of clients. \textbf{B:} The clients train the model locally using their data. \textbf{C:} The local models are sent to an aggregation server. \textbf{D:} The client models are aggregated to make a new global model. This process is repeated until convergence.}
    \label{fig:federatedLearning}
\end{figure}

\section{Robust aggregation methods}

\begin{table}[ht]
    \centering
    \begin{tabular}{l|m{7.0cm}|c|c}
         Name & Description  & Robust & Server data   \\ \hline \hline
         FA (or FedAvg)   & Federated Averaging \cite{mcmahan2017communication} & No & No  \\ \hline
         COMED    & Coordinate-wise median \cite{yin2018byzantine} & Yes & No  \\ \hline
         MKRUM    & Multi-Krum \cite{blanchard2017machine} & Yes & No  \\ \hline
         AFA      & Adaptive Federated Averaging \cite{munoz2019byzantine} & Yes & No  \\ \hline
         FedMGDA++ &  FedMGDA+ \cite{hu2020fedmgda+}  & Yes & No  \\ \hline
         FedDF    & FedDF \cite{lin2020ensembleFedDF} & No & Yes \\ \hline
         \textbf{FedDFmed}    & FedDF with median-based Knowledge Distillation & Yes & Yes \\ \hline
         FedBE    & FedBE \cite{chen2020fedbe} & No & Yes \\ \hline
         \textbf{FedBEmed}    & FedBE with median-based Knowledge Distillation & Yes & Yes \\ \hline
         \textbf{FedADF}   & AFA \cite{munoz2019byzantine} with an additional median-based Knowledge Distillation & Yes & Yes \\ \hline
         \textbf{FedMGDA+DF}   & FedMGDA+ \cite{hu2020fedmgda+} with an additional median-based Knowledge Distillation & Yes & Yes \\ \hline
         \textbf{FedRAD}   & Our novel aggregator, see Algorithm \ref{alg:FedRAD}  & Yes & Yes \\ \hline
         \textbf{FedRADnoise}   & FedRAD using uniform noise instead of server-side data & Yes & No \\ \hline
    \end{tabular}
    \caption[List of Aggregation Methods]{List of aggregation methods. Methods in \textbf{bold} are novel. "Robust" is used to describe whether steps have been taken to detect or defend against adversaries. "Server data" is used to indicate whether an \emph{unlabelled} dataset is required on the server side.}
    \label{tab:aggregators}
\end{table}






\section{Robustness of the median}\label{app:medianScores}

Table \ref{tab:medianRobust} illustrates the advantage of using a robust statistic such as the median instead of mean for model aggregation.

\begin{table}[ht]
    \centering
    \begin{tabular}{|r|c|c|c|}
    \hline 
         & Logits predictions from 10 clients & Average & Median \\ \hline 
         No attacker & $[1,1,2,2,3,3,4,4,5,5]$ & $3$ & $3$ \\ 
         One weak attacker & $[1,1,2,2,3,3,4,4,5,\textcolor{red}{15}]$ & $4$ & $3$ \\ 
         Four weak attackers & $[1,1,2,2,3,3,\textcolor{red}{14},\textcolor{red}{14},\textcolor{red}{15},\textcolor{red}{15}]$ & $7$ & $3$ \\ 
         One strong attackers & $[1,1,2,2,3,3,4,4,5,\textcolor{red}{1005}]$ & $103$ & $3$ \\ 
         Four strong attackers & $[1,1,2,2,3,3,\textcolor{red}{1004},\textcolor{red}{1004},\textcolor{red}{1005},\textcolor{red}{1005}]$ & $403$ & $3$ \\ \hline 
    \end{tabular}
    \caption[Robustness of Median]{An illustrative example demonstrating the robustness of the median against attackers with confidently incorrect predictions. The logits values are hypothetical outputs from 10 different models for a particular class. Attackers are coloured in red. }
    \label{tab:medianRobust}
\end{table}


\section{Learning curves of attacks}\label{app:learningCurves}
\subsection{No attacks}
\begin{figure}[ht] 
\centering
    \adjustbox{max width=\textwidth}{
        \includegraphics[trim={95 250 95 250}, clip=True]{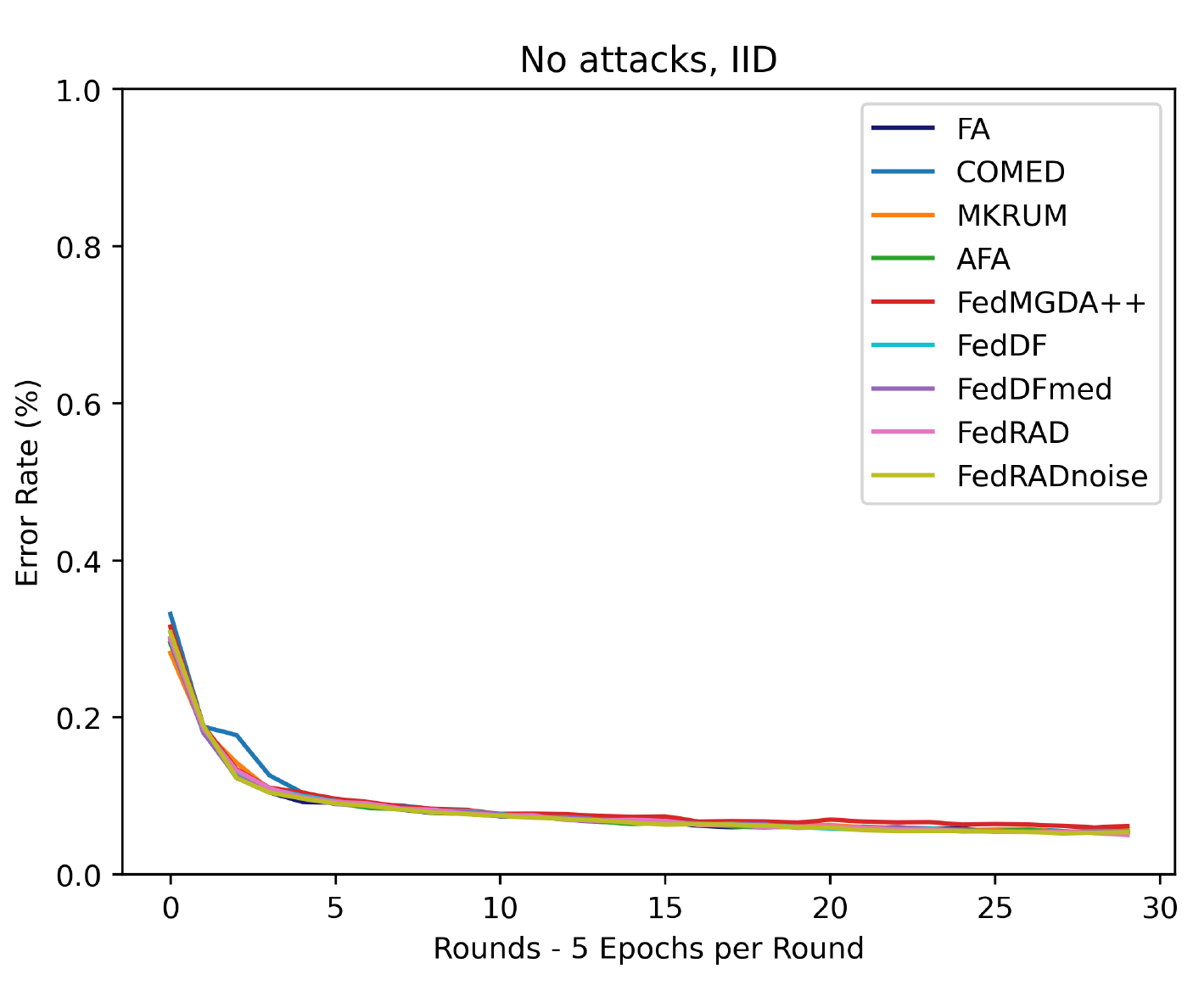}
        \includegraphics[trim={95 250 95 250}, clip=True]{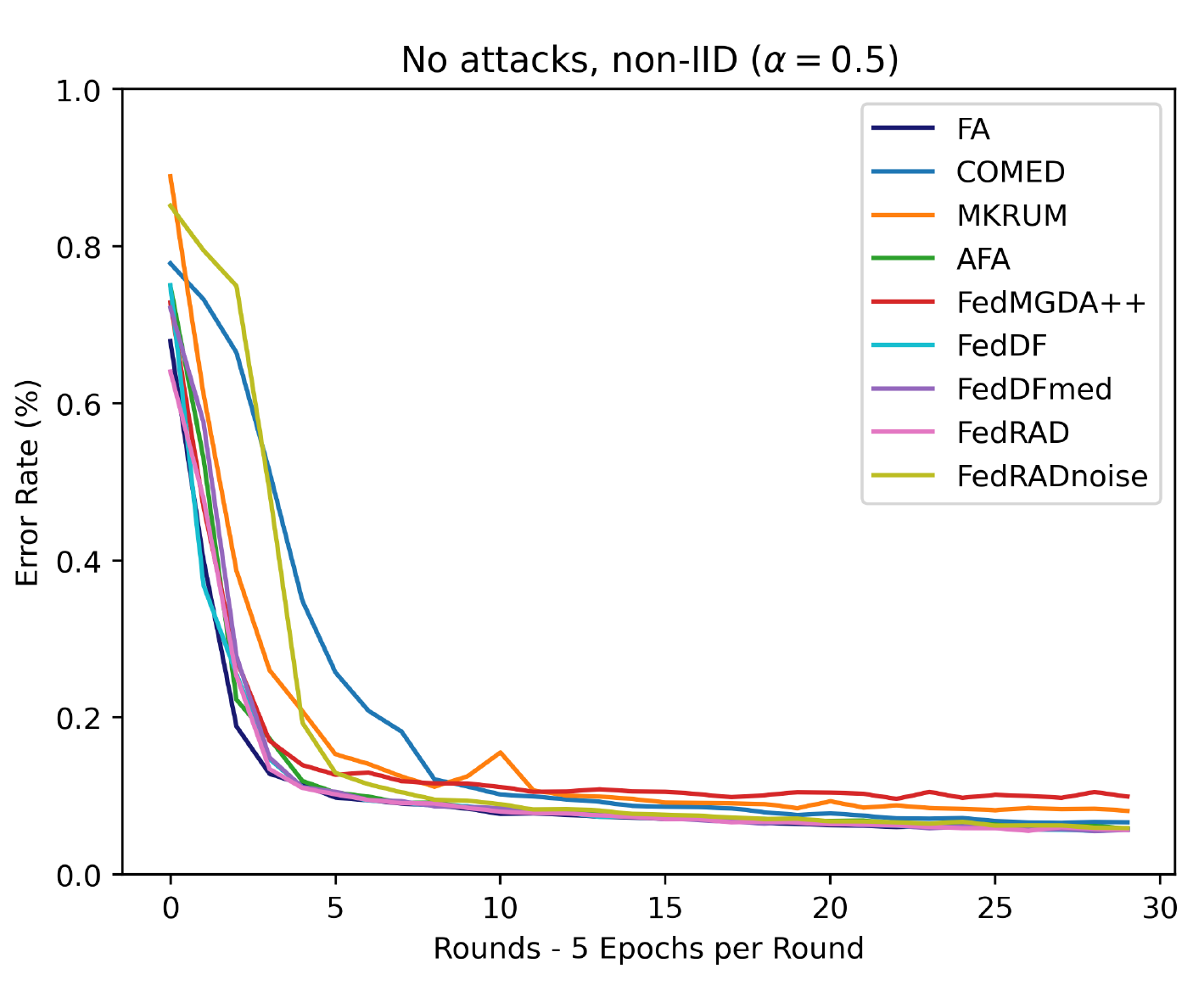}
        \includegraphics[trim={95 250 95 250}, clip=True]{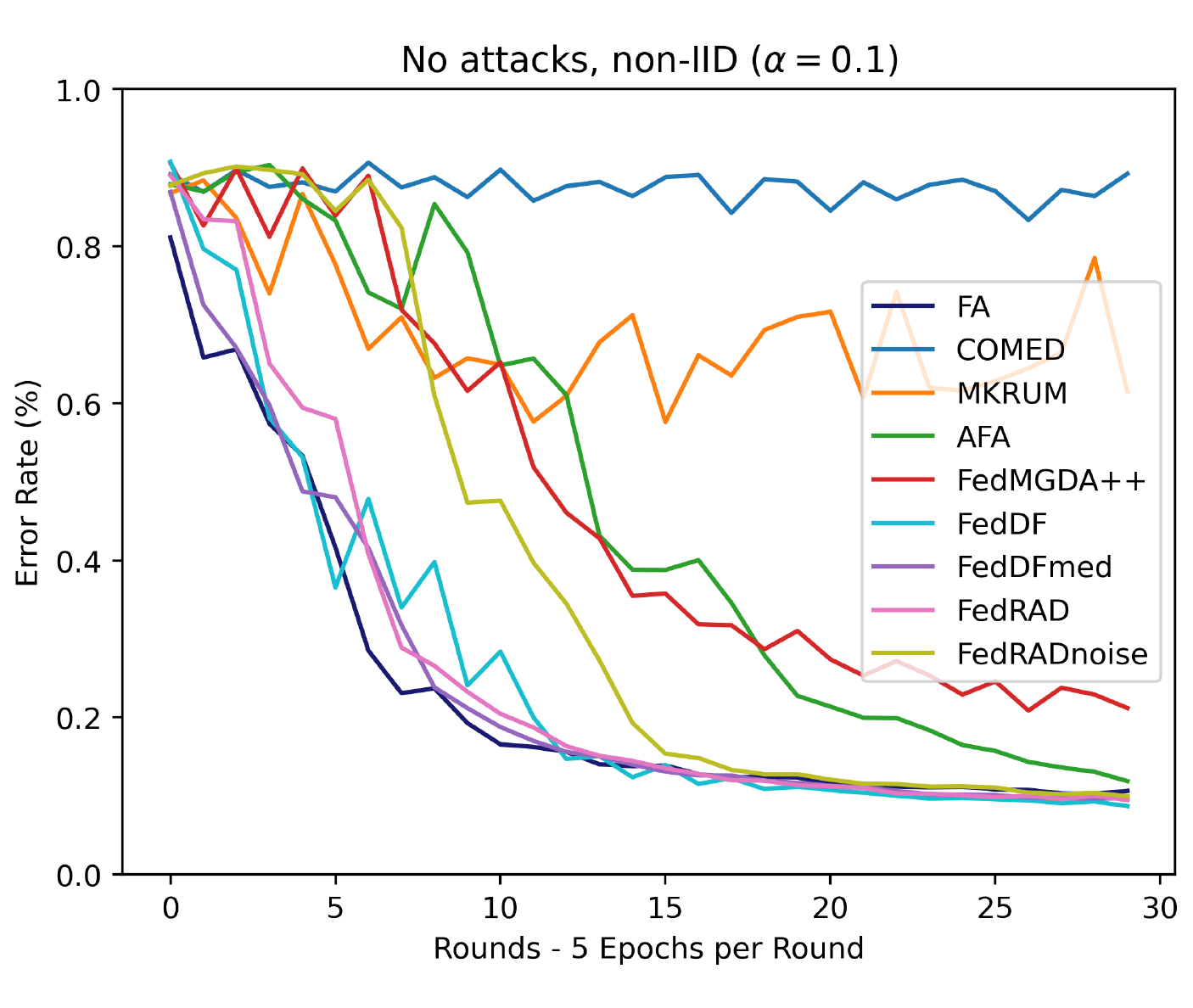}
    }
\caption{Effects of non-IID data for no attacks.}
\label{fig:no_attack_non-iid}
\end{figure}

Results from experiments with no attacks for different levels of data heterogeneity can be seen in Figure \ref{fig:no_attack_non-iid}. 

All the methods perform roughly the same on IID data when there are no attackers.
With increased heterogeneity, the learning starts to slow down for all aggregators. When data is very non-IID, i.e. $\alpha=0.1$, the performance of aggregators such as COMED, MKRUM, AFA and FedMGDA+ declines significantly. Both FedMGDA+ and AFA slow down quite a bit, and both of them block some healthy clients which effectively decreases the size of the training set.
 
Federated Averaging still performs well in non-IID circumstances when there are no attackers. Methods that use Knowledge Distillation also perform similarly to FedAvg.

\subsection{Faulty Attacks}
\begin{figure}[ht] 
\centering
    \adjustbox{max width=\textwidth}{
        \includegraphics[trim={95 250 95 250}, clip=True]{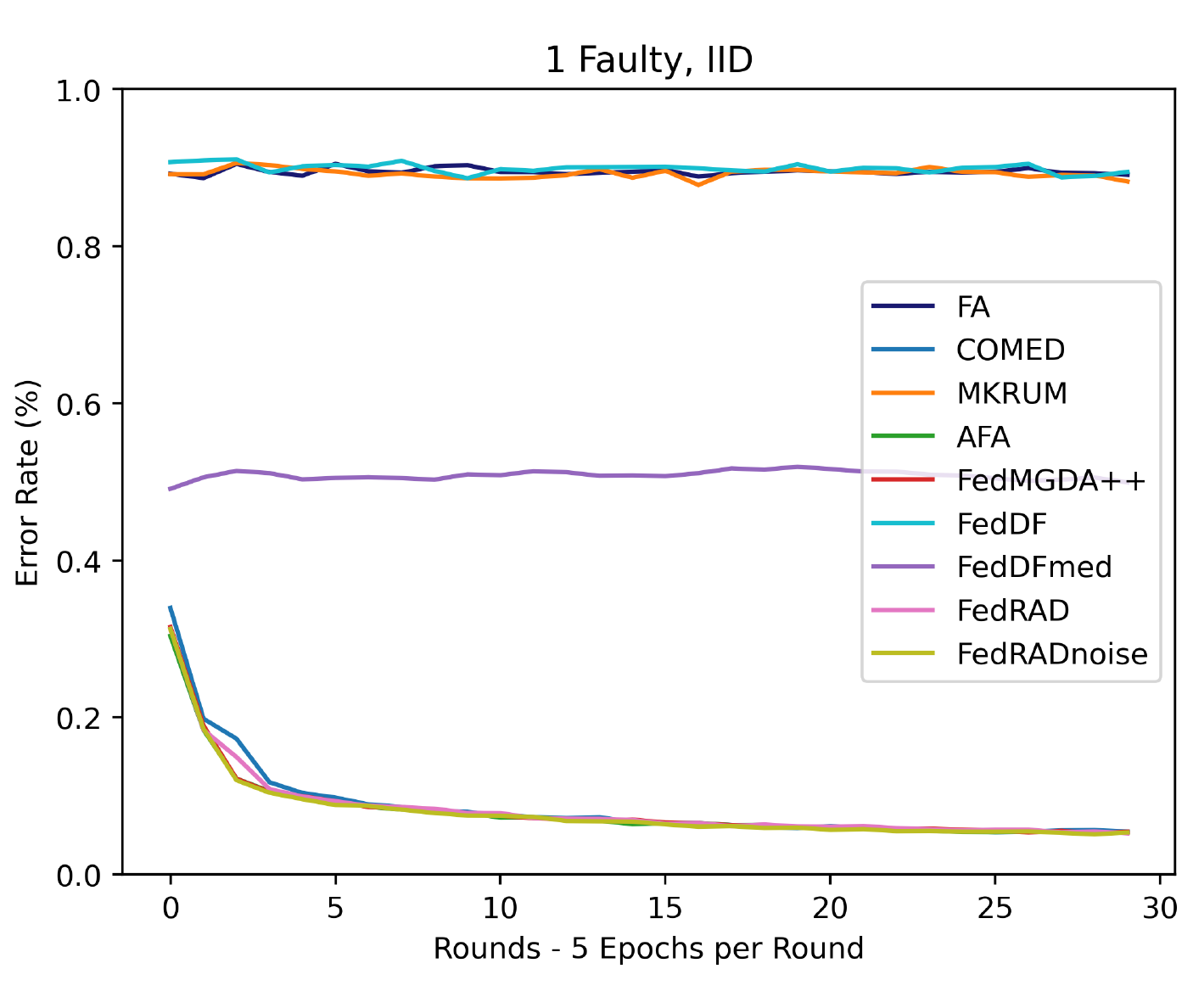}
        \includegraphics[trim={95 250 95 250}, clip=True]{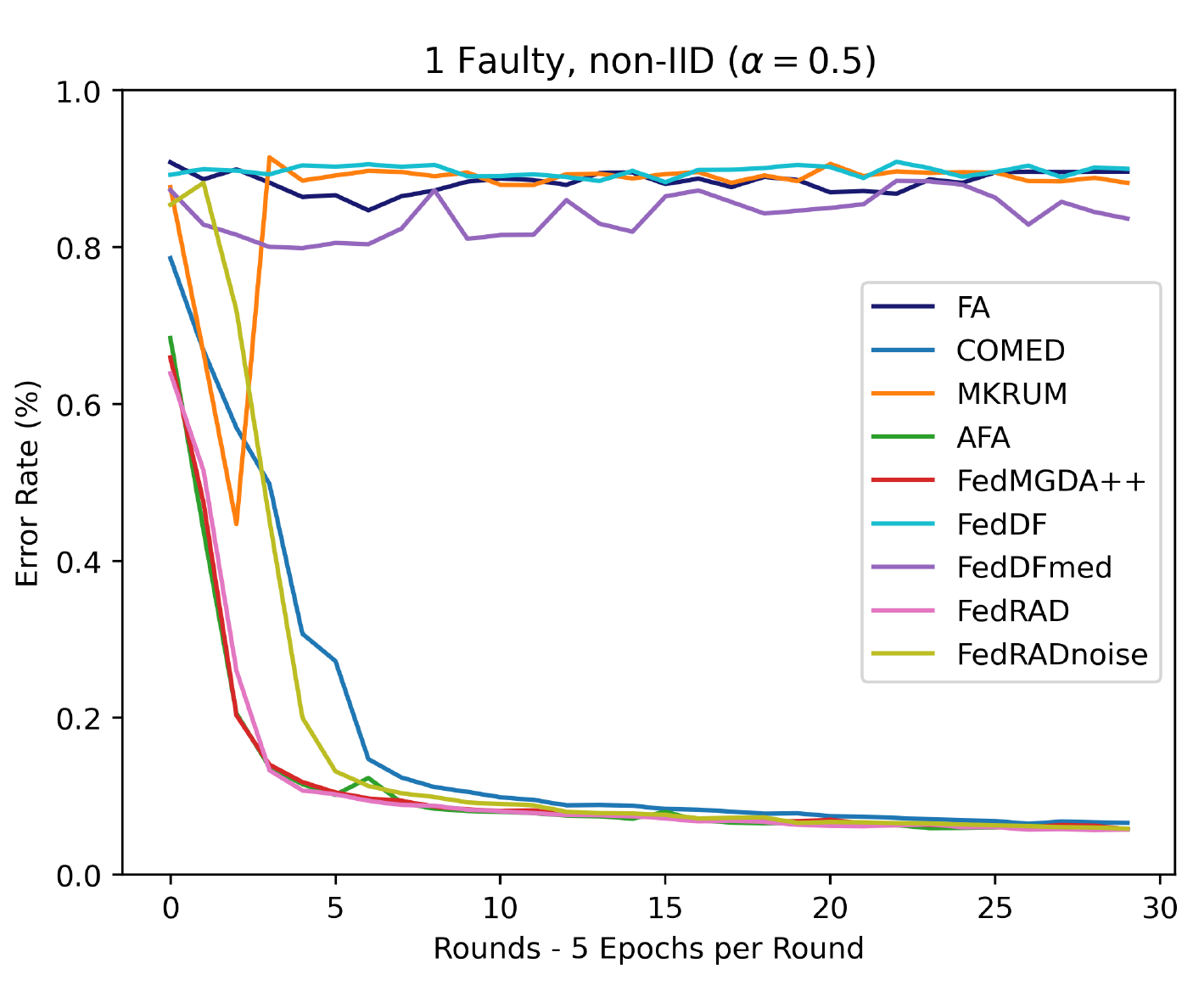}
        \includegraphics[trim={95 250 95 250}, clip=True]{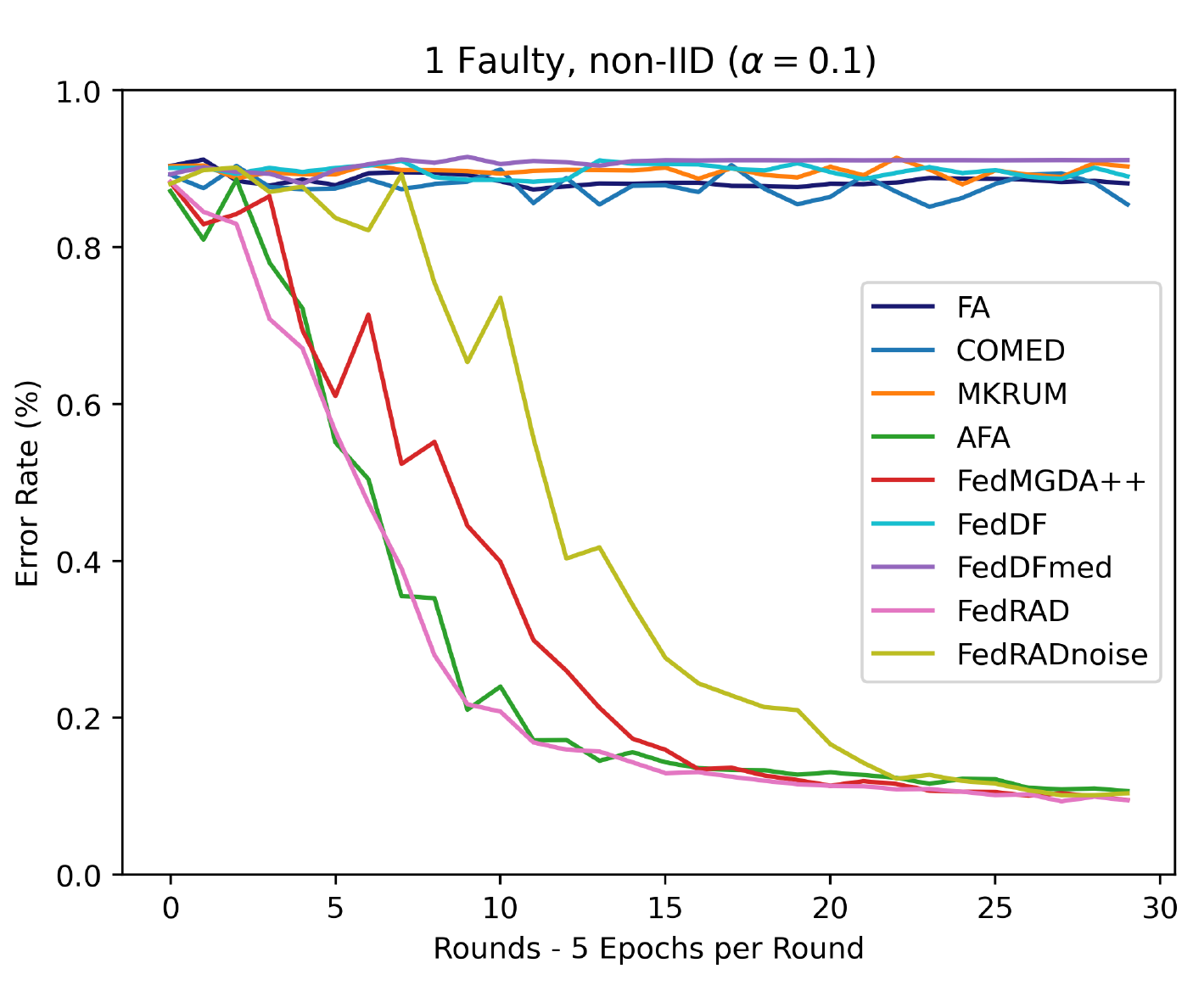}
    }
\caption{Effects of non-IID data with $1$ Faulty attacker.}
\label{fig:attack_non-iid_faulty1}
\end{figure}

\begin{figure}[ht] 
\centering    
    \adjustbox{max width=\textwidth}{
        \includegraphics[trim={95 250 95 250}, clip=True]{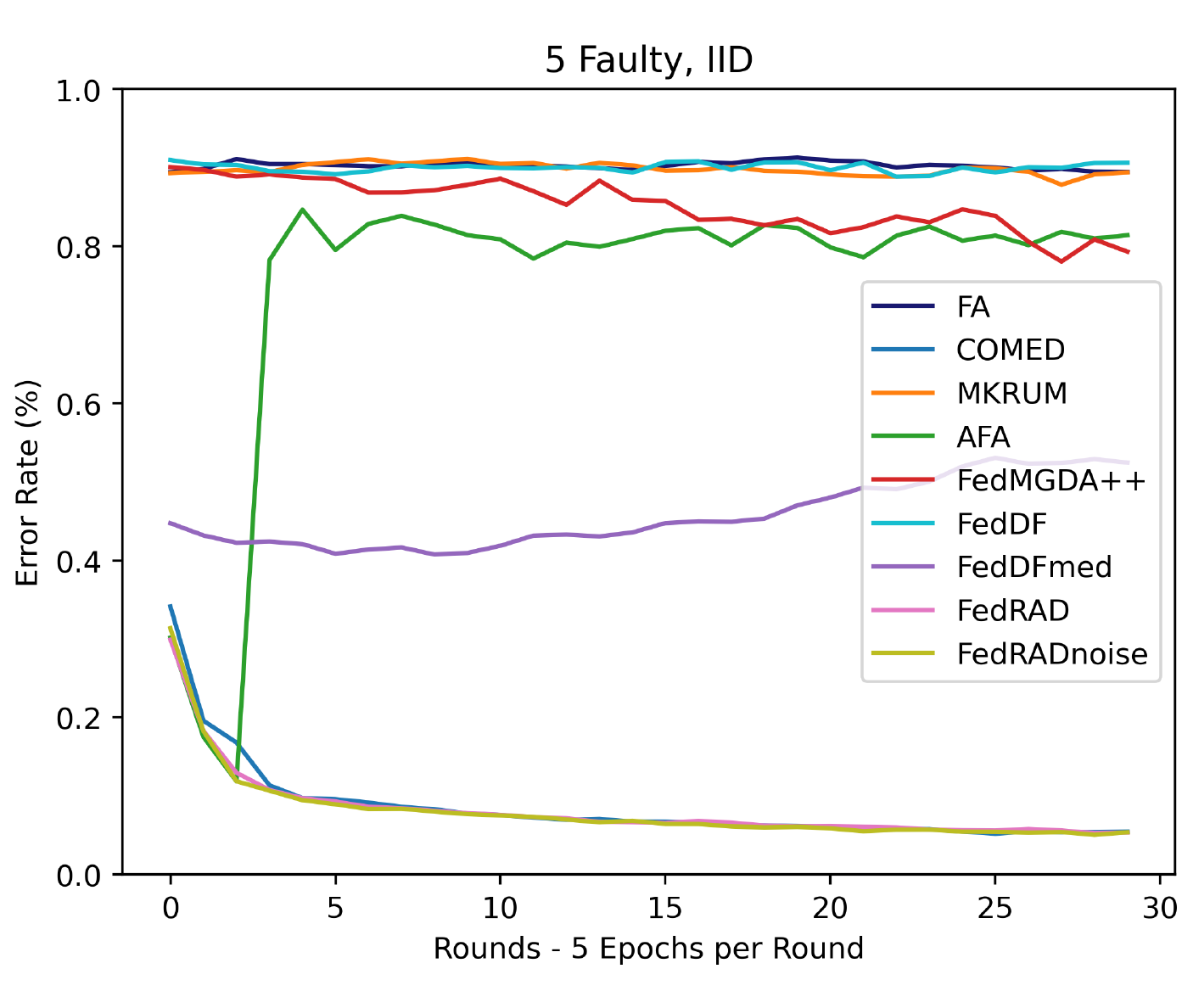}
        \includegraphics[trim={95 250 95 250}, clip=True]{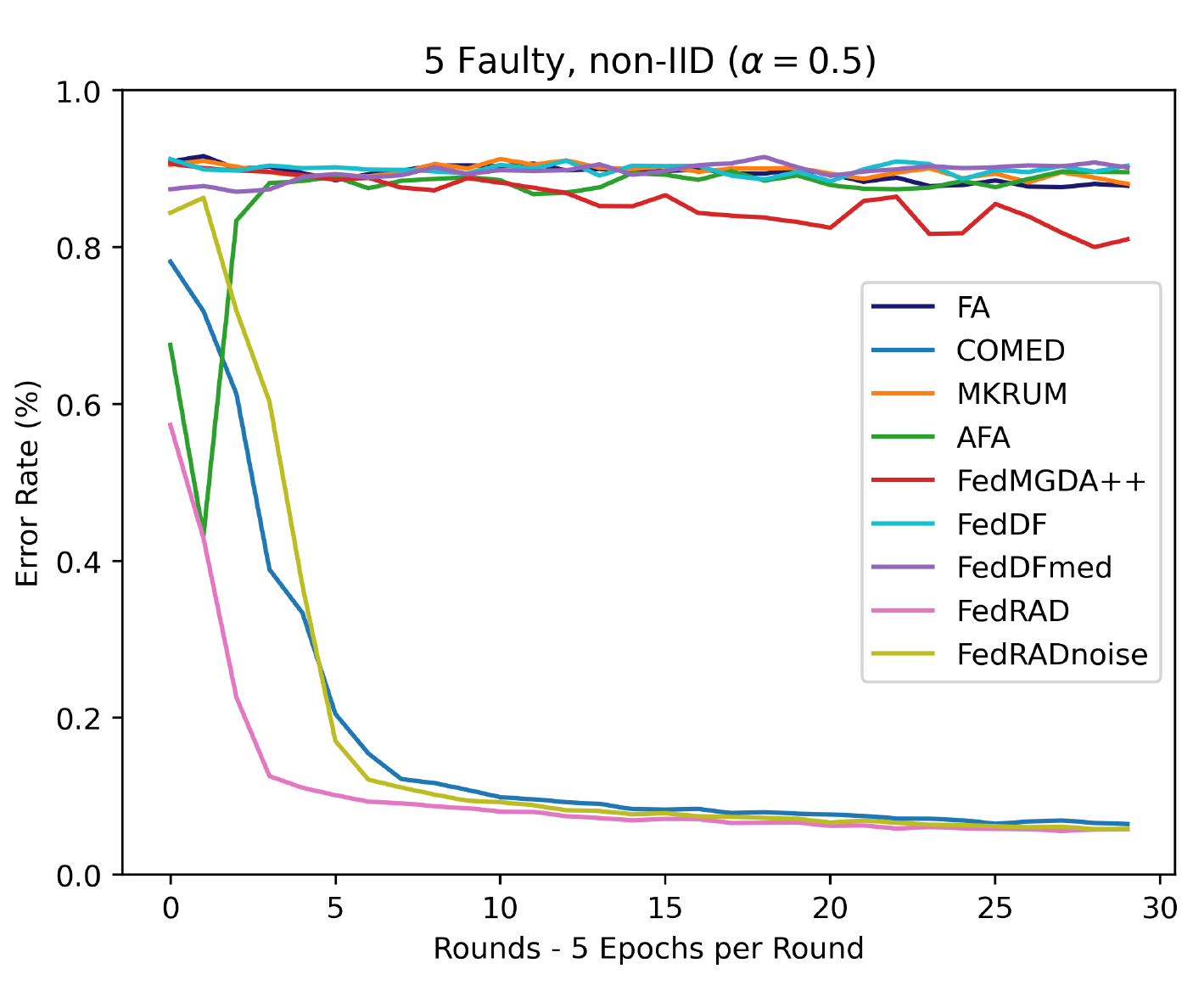}
        \includegraphics[trim={95 250 95 250}, clip=True]{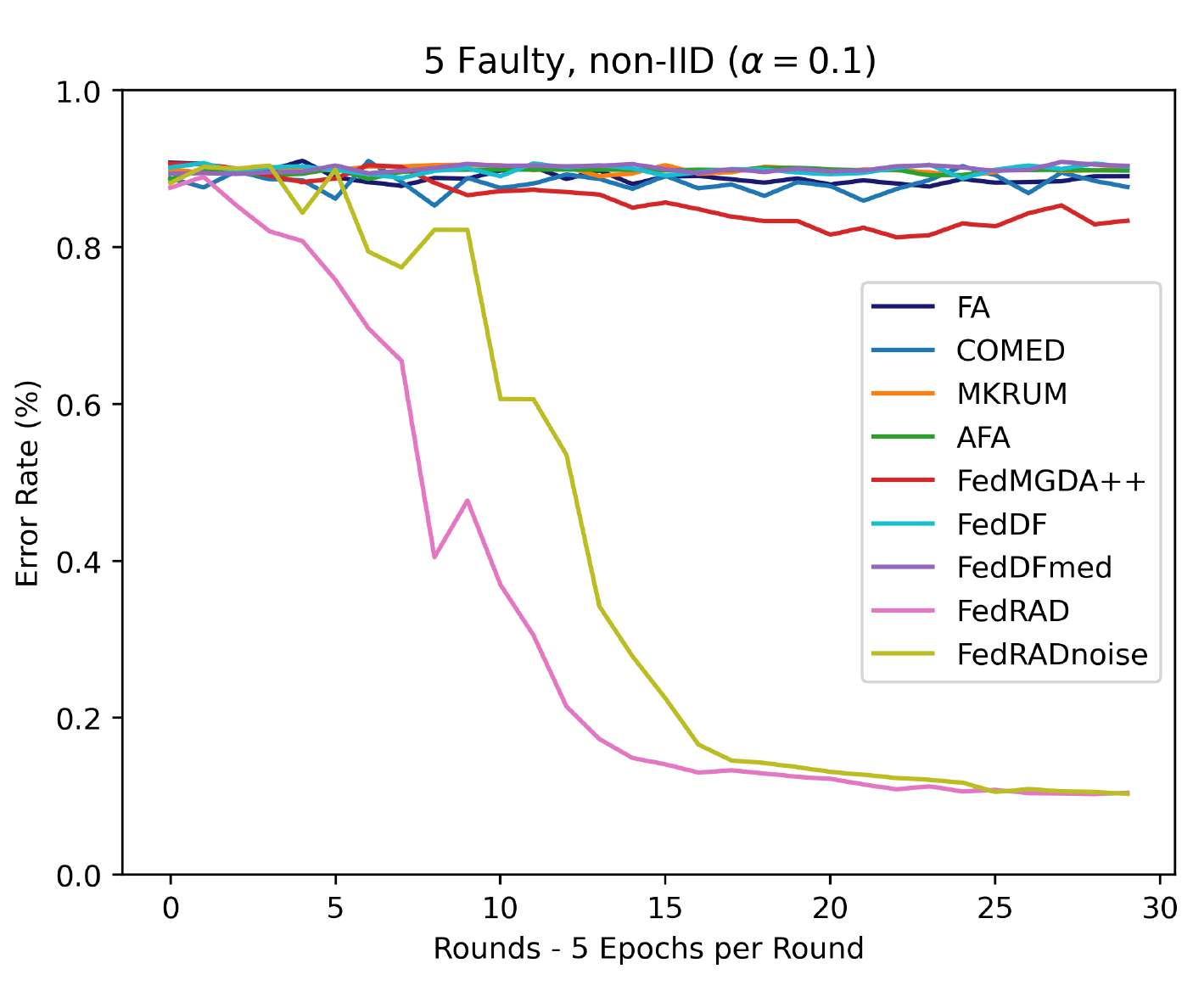}
    }
\caption{Effects of non-IID data with $5$ Faulty attackers.}
\label{fig:attack_non-iid_faulty5}
\end{figure}

\begin{figure}[ht] 
\centering
    \adjustbox{max width=\textwidth}{
        \includegraphics[trim={95 250 95 250}, clip=True]{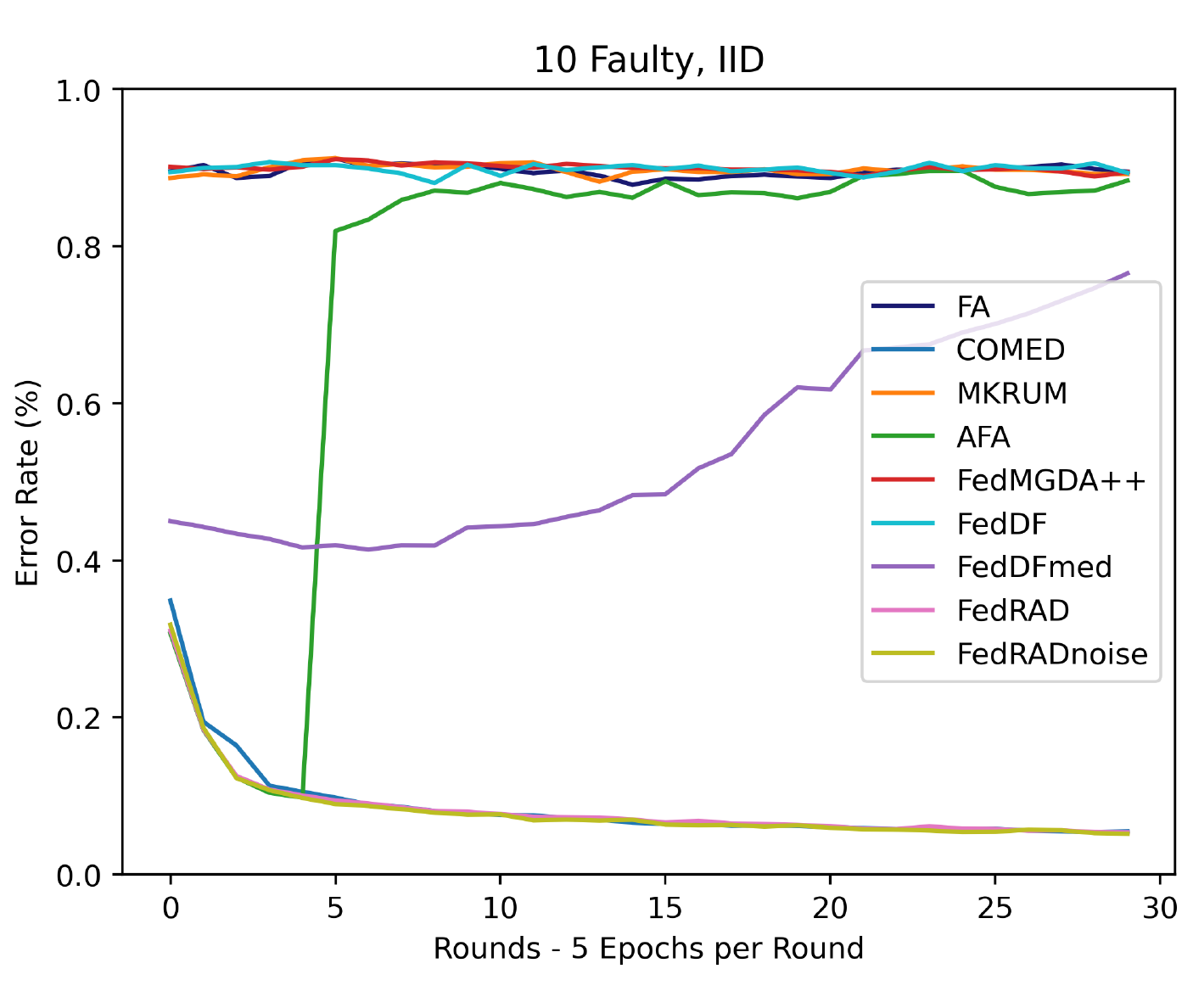}
        \includegraphics[trim={95 250 95 250}, clip=True]{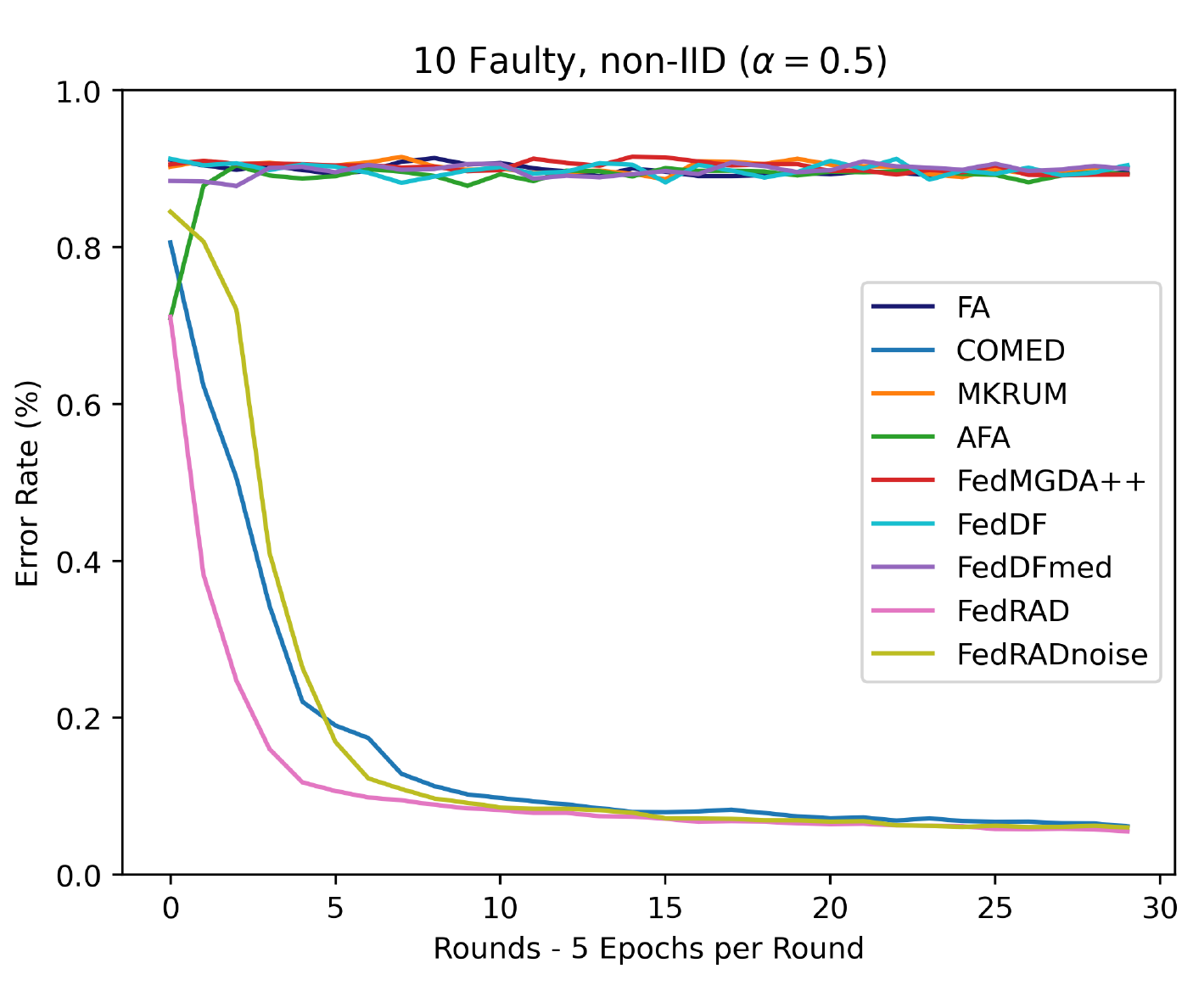}
        \includegraphics[trim={95 250 95 250}, clip=True]{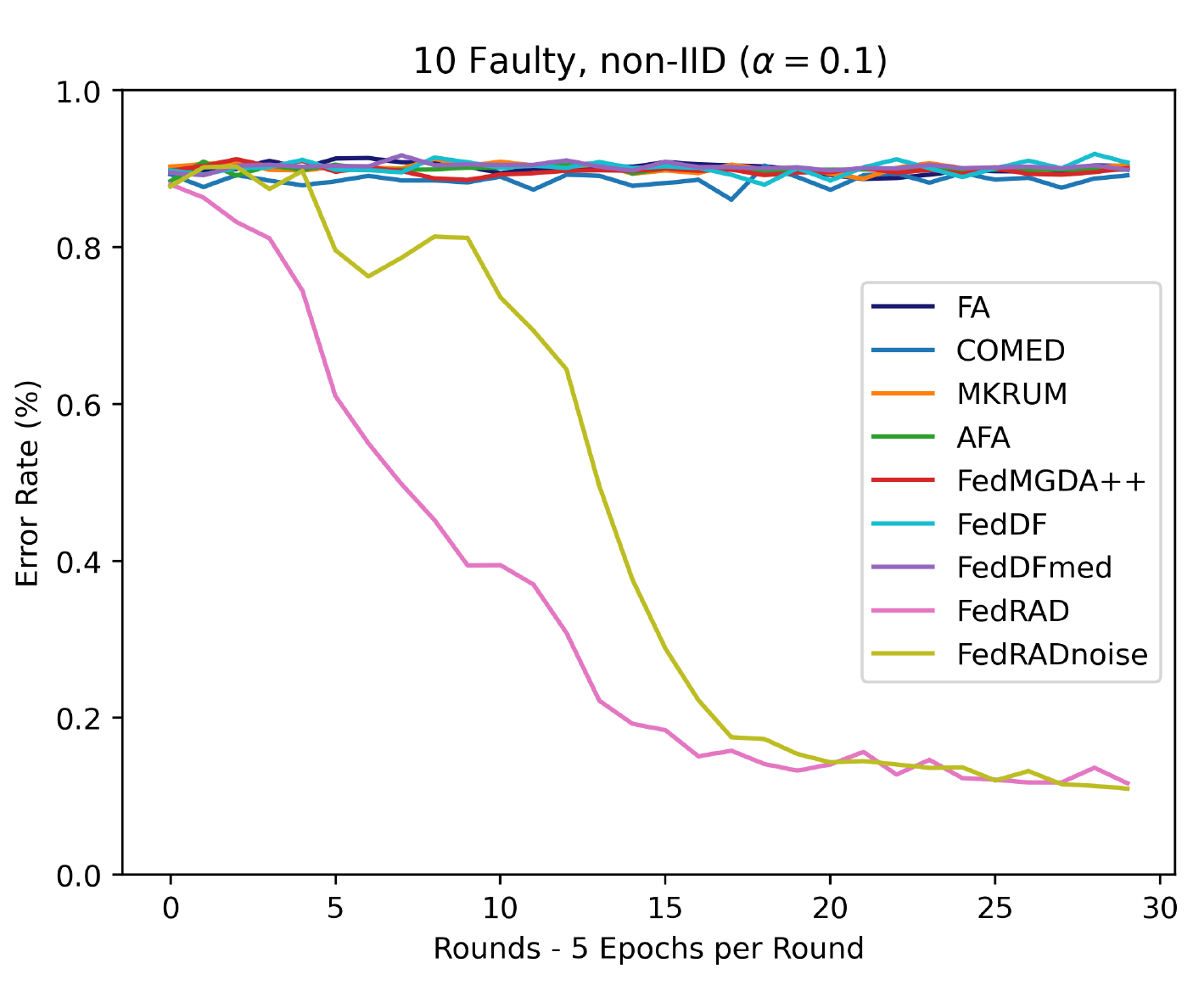}
    }
\caption{Effects of non-IID data with $10$ Faulty attackers.}
\label{fig:attack_non-iid_faulty10}
\end{figure}

Faulty attackers, also known as Byzantine, are attackers which update their models by adding a lot of random noise to the model parameters. Results for 1, 5 and 10 faulty attackers can be seen in Figures \ref{fig:attack_non-iid_faulty1}, \ref{fig:attack_non-iid_faulty5} and \ref{fig:attack_non-iid_faulty10} respectively.

Just $1$ faulty attacker completely ruins performance for non-robust aggregators. Both AFA and FedMGDA+ are able to effectively block the attacker and learn, even with non-IID data. 

Once more attackers are added, AFA and FedMGDA+ start to fail. COMED still works for homogeneous data. The only aggregators that work against 5 faulty clients in the most heterogeneous case are our novel methods.

With 10 attackers and non-IID data FedRAD is the only aggregator that manages to learn anything. Even FedRADnoise, which uses only random noise for scoring and distillation, is able to learn well.
The good performance of aggregators which use our novel median-scoring mechanism against is explained by Figure \ref{fig:median-histograms}, which shows that Faulty models rarely ever give the median logit prediction for a given class, and thus often get a score of $0$, which is equivalent to leaving faulty models out completely during the weighted averaging.

\subsection{Malicious Attacks}
\begin{figure}[ht] 
\centering
    \adjustbox{max width=\textwidth}{
        \includegraphics[trim={95 250 95 250}, clip=True]{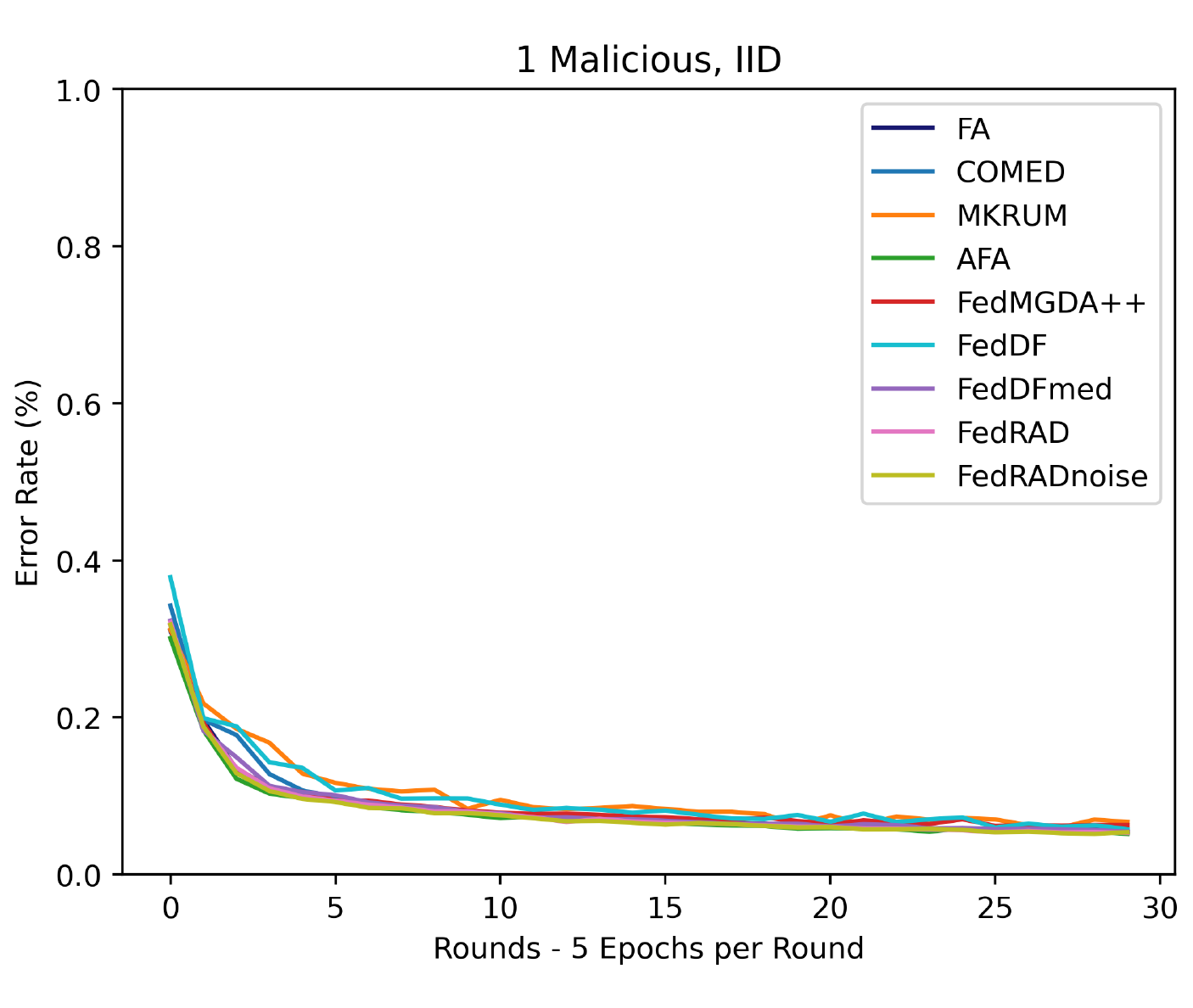}
        \includegraphics[trim={95 250 95 250}, clip=True]{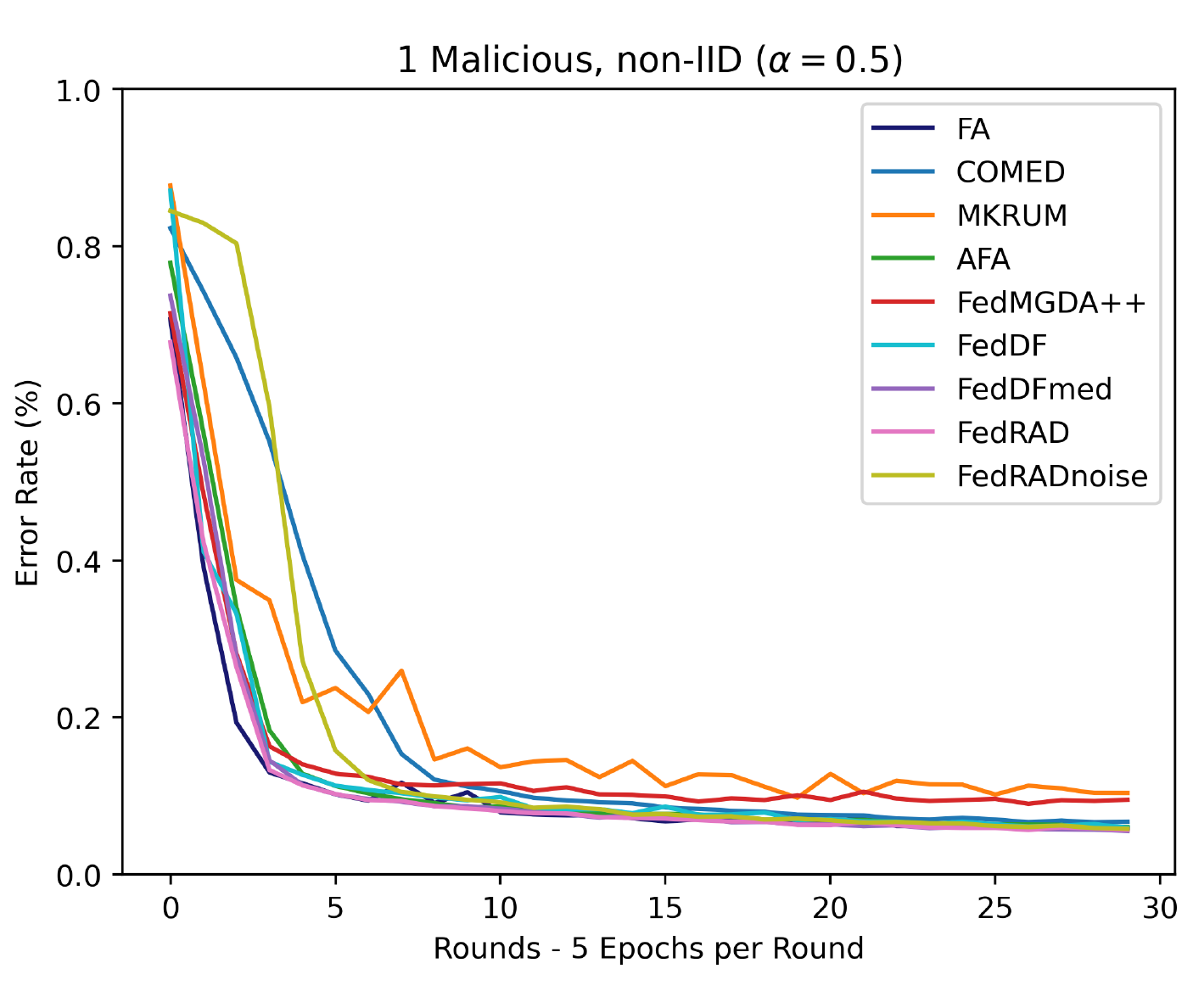}
        \includegraphics[trim={95 250 95 250}, clip=True]{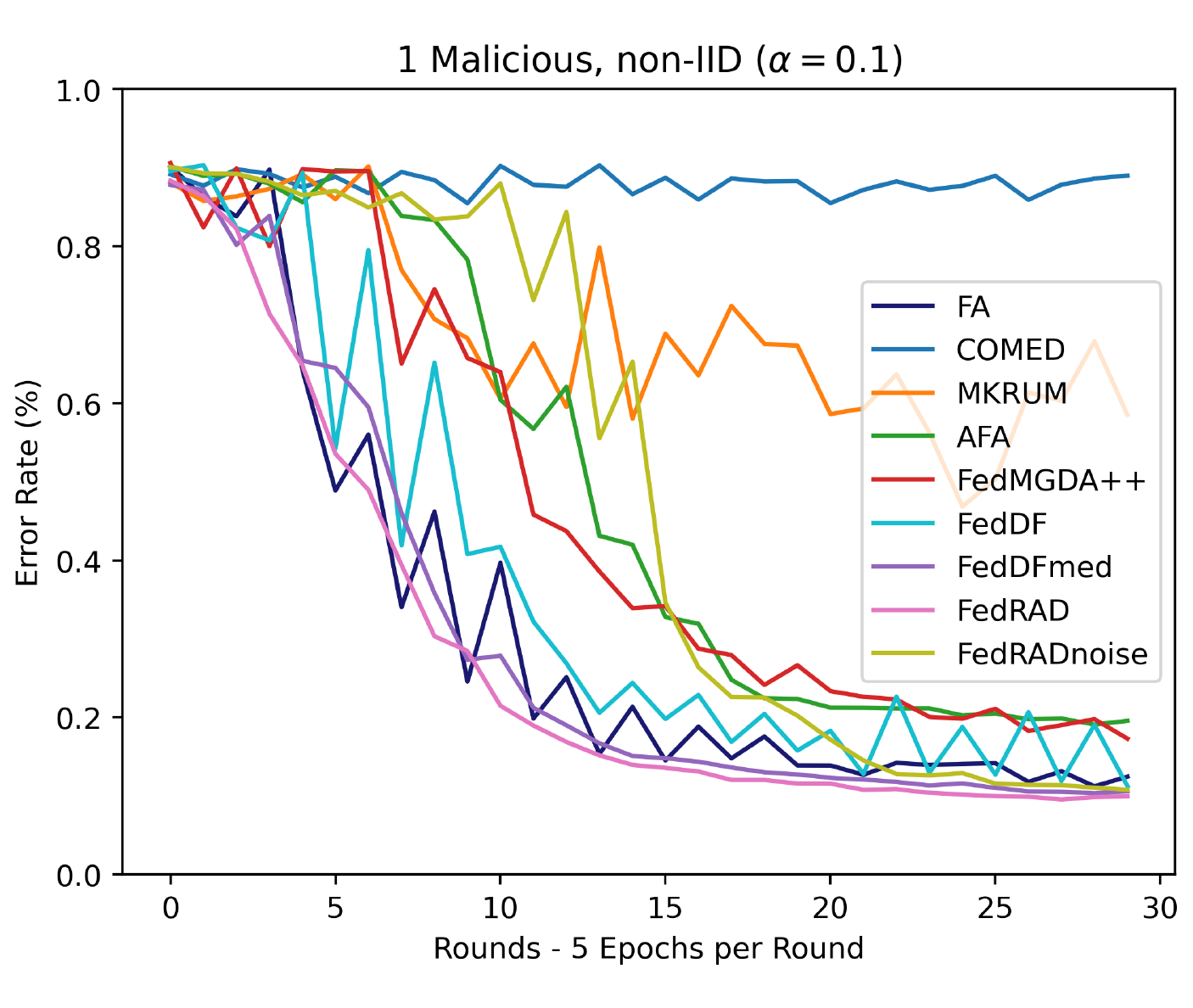}
    }
\caption{Effects of non-IID data with $1$ malicious attacker.}
\label{fig:attack_non-iid_mal1}
\end{figure}

\begin{figure}[ht] 
\centering
    \adjustbox{max width=\textwidth}{
        \includegraphics[trim={95 250 95 250}, clip=True]{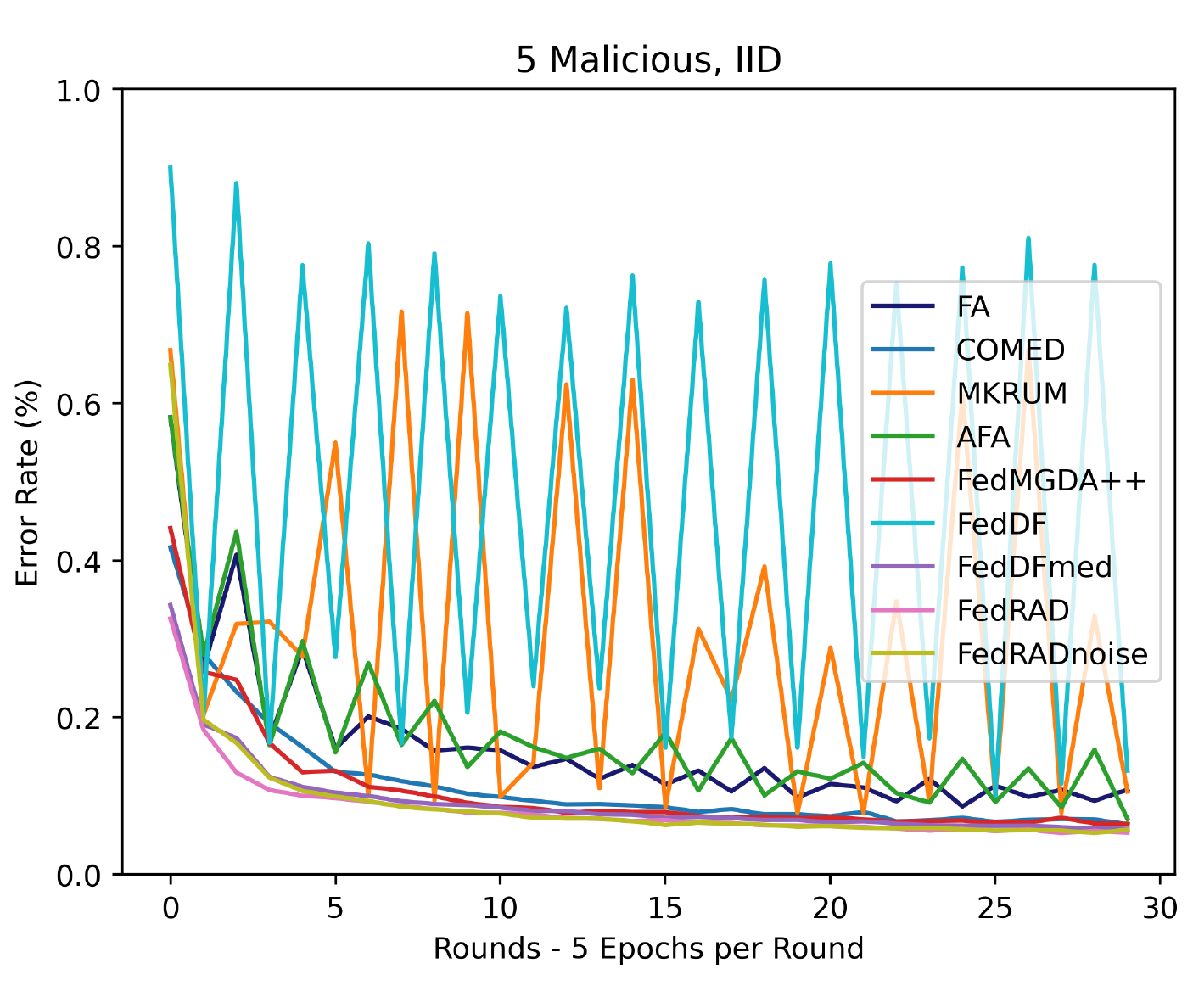}
        \includegraphics[trim={95 250 95 250}, clip=True]{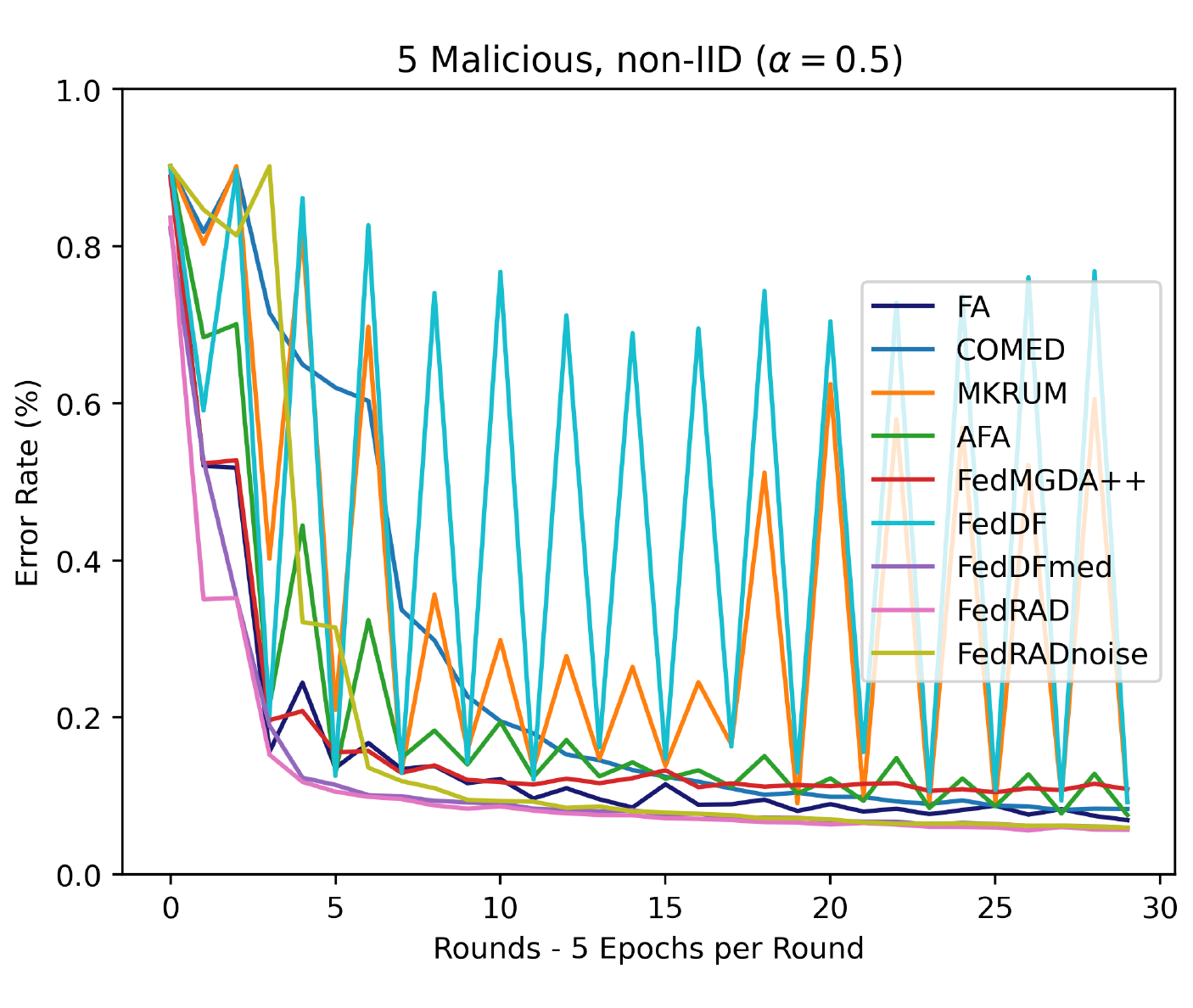}
        \includegraphics[trim={95 250 95 250}, clip=True]{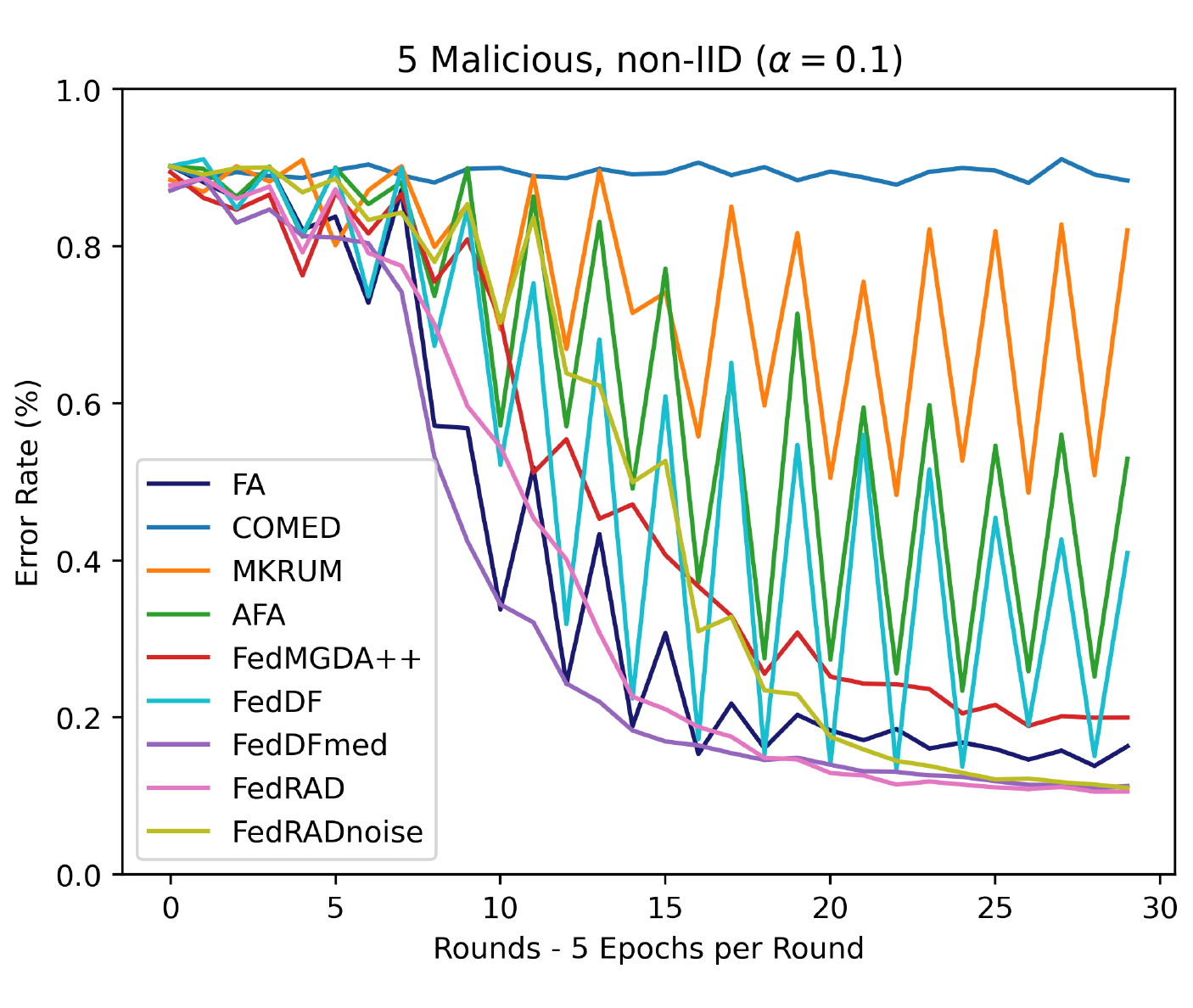}
    }
\caption{Effects of non-IID data with $5$ malicious attackers.}
\label{fig:attack_non-iid_mal5}
\end{figure}

\begin{figure}[ht] 
\centering
    \adjustbox{max width=\textwidth}{
        \includegraphics[trim={95 250 95 250}, clip=True]{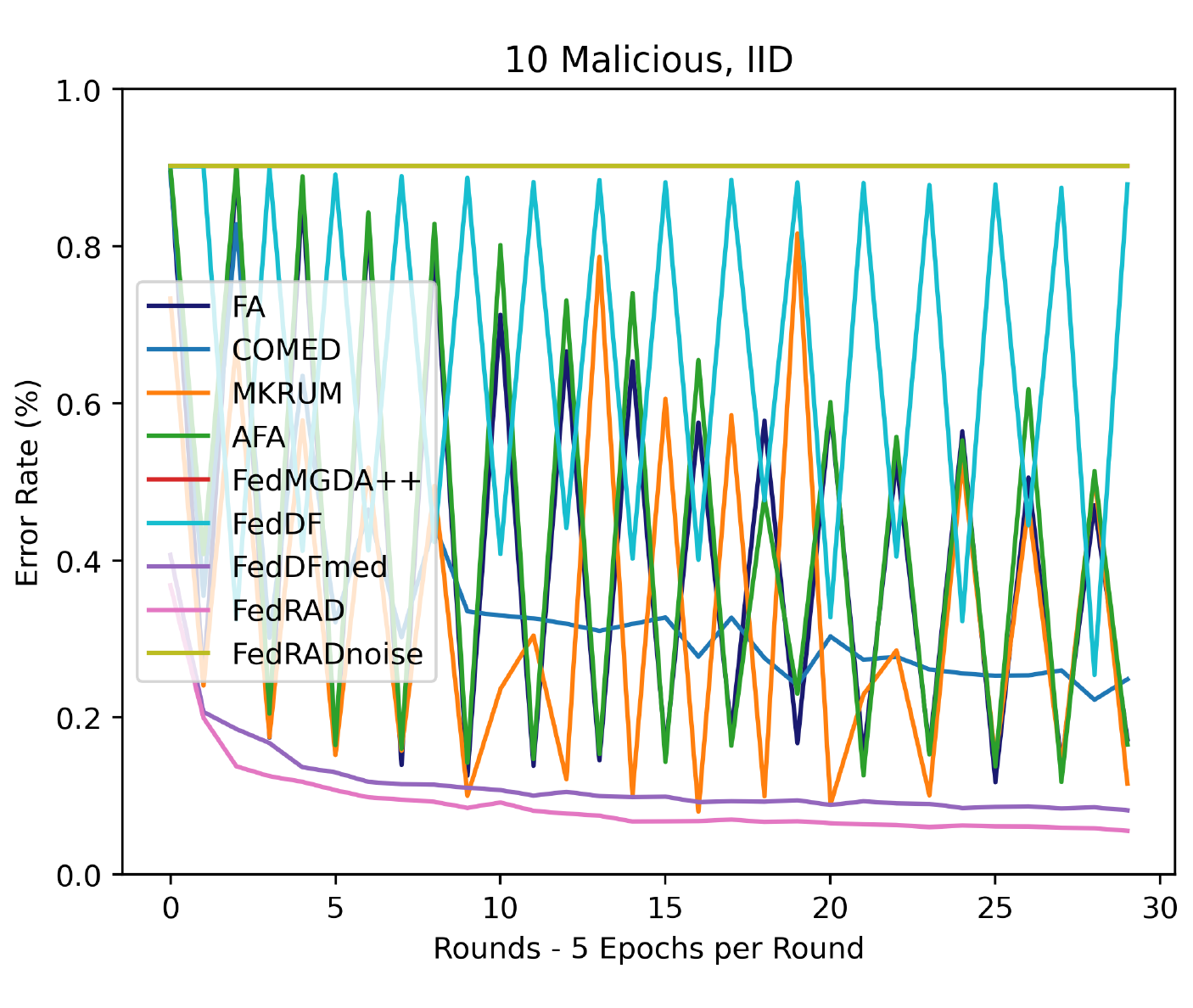}
        \includegraphics[trim={95 250 95 250}, clip=True]{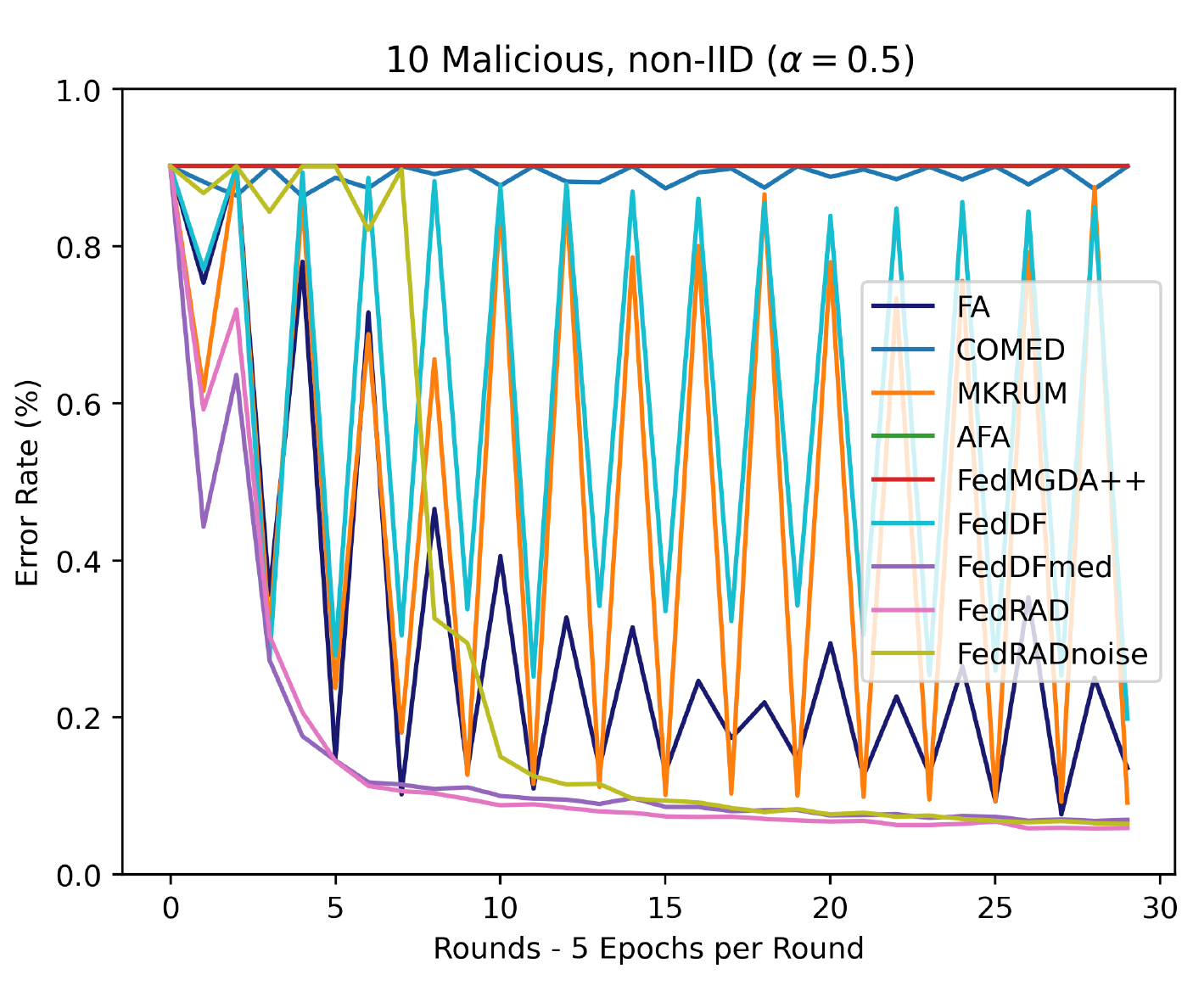}
        \includegraphics[trim={95 250 95 250}, clip=True]{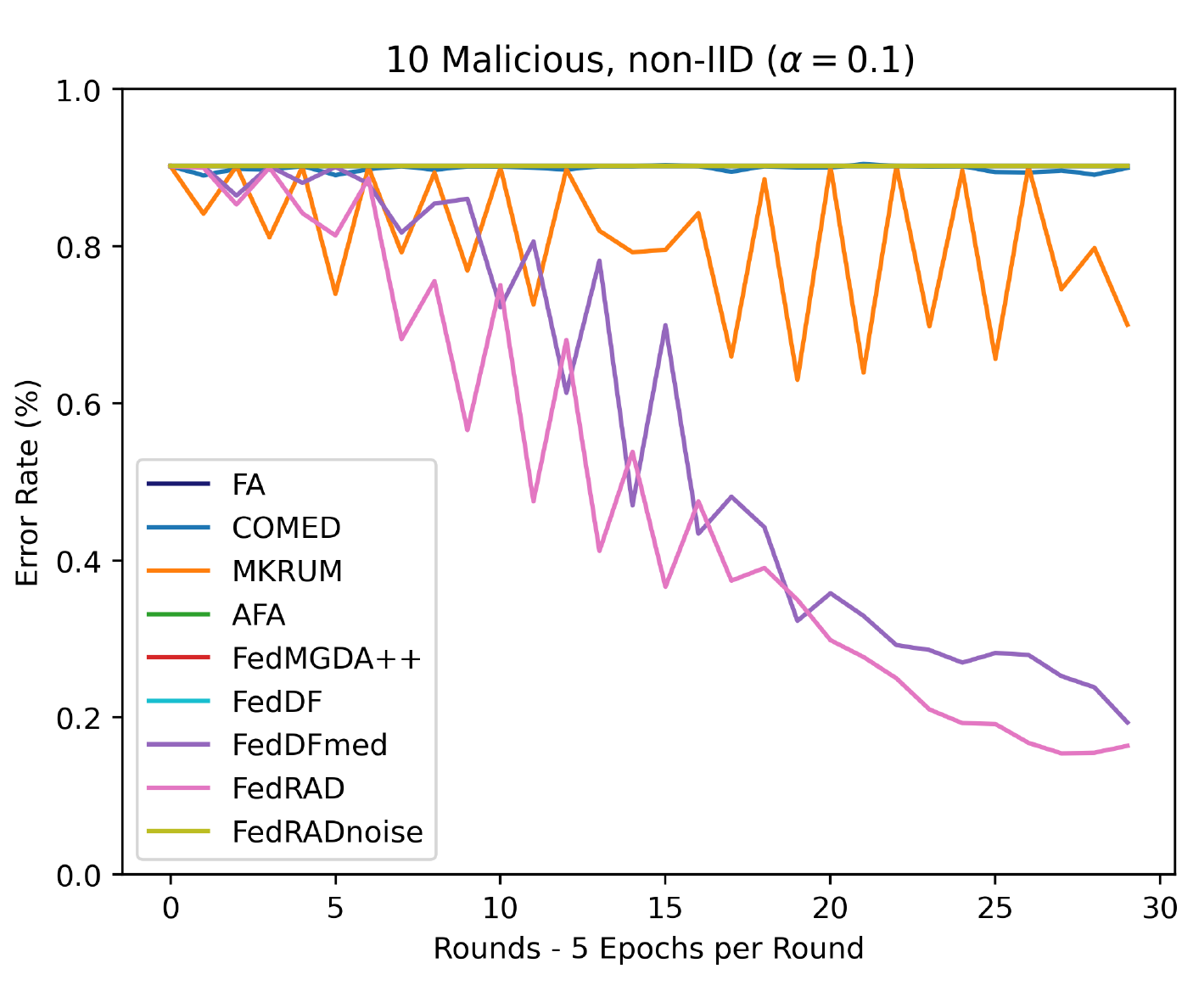}
    }
\caption{Effects of non-IID data with $10$ malicious attackers.}
\label{fig:attack_non-iid_mal10}
\end{figure}

Malicious attackers are agents who train the models with incorrect labels. In our case, all labels are set to $0$ before training. Results for 1, 5, and 10 attackers can be seen in Figures \ref{fig:attack_non-iid_mal1}, \ref{fig:attack_non-iid_mal5} and \ref{fig:attack_non-iid_mal10} respectively. 

One malicious attacker does not impact learning very much. 

With 5 malicious attackers, many aggregators start to perform badly and the training becomes very noisy. 
This is where the benefits of median-based Knowledge distillation starts to shine through: FedDF training is very noisy, but FedDFmed is among the best performing methods.

With 10 malicious attackers, even more aggregators start to fail. Robust methods which rely on comparing model parameters, such as AFA and FedMGDA+, don't learn anything at all in the non-IID case.

The best performing aggregators against malicious attacks in both IID and non-IID scenarios are our novel methods which utilize median-based Knowledge Distillation: FedDFmed and FedRAD. This is explained by the median-counting histograms in Figure \ref{fig:median-histograms}, which show that median logits are less contaminated by malicious agents.

MKRUM, AFA and FedMGDA+ fail because models from malicious agents are quite similar to healthy models when comparing their high-dimensional parameters. In non-IID situations, it becomes difficult to detect malicious models among the non-IID healthy models by using distance metrics on their parameters.

\subsection{Both Types of Attacks}
\begin{figure}[ht] 
\centering
    \adjustbox{max width=\textwidth}{
        \includegraphics[trim={95 250 95 250}, clip=True]{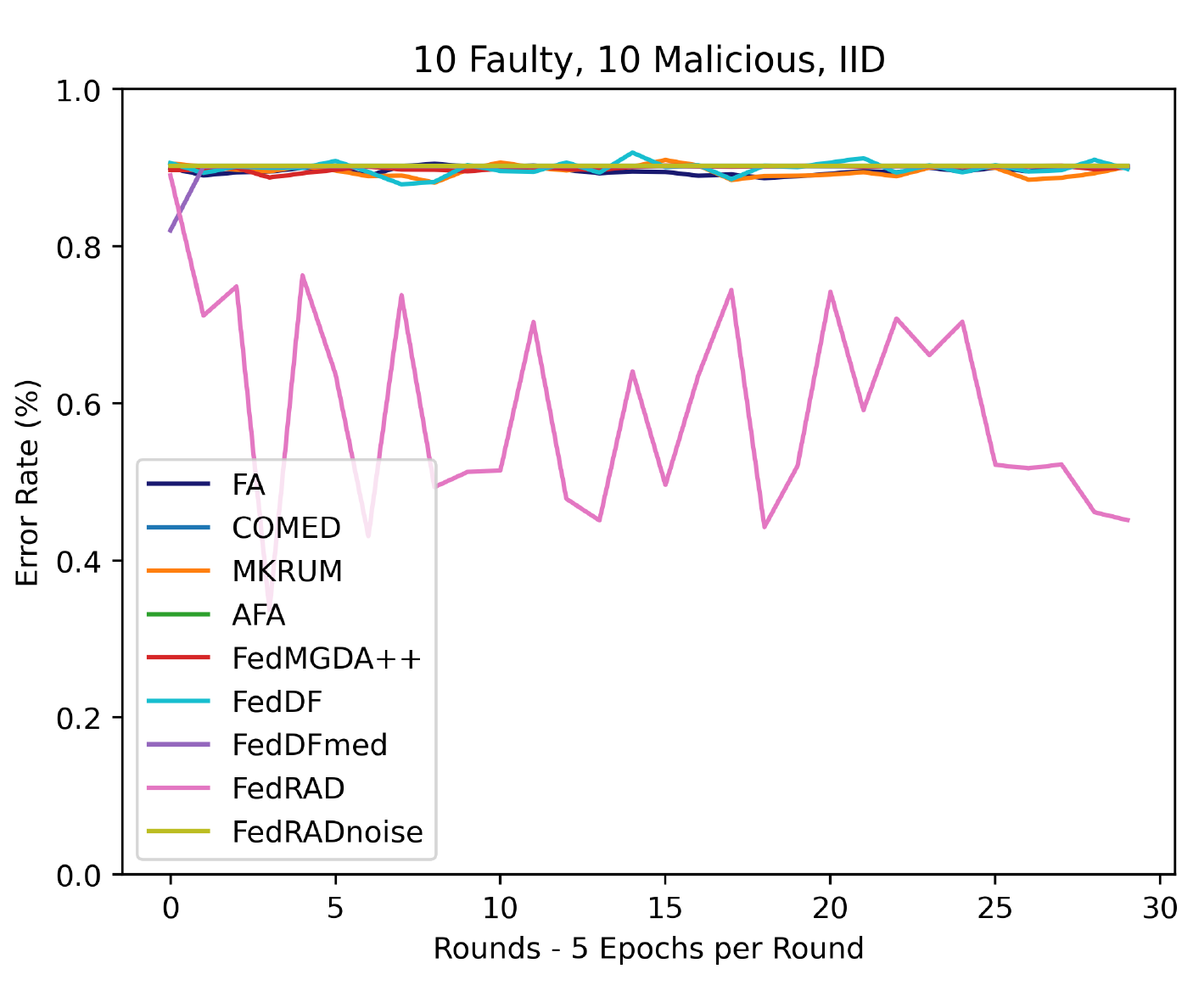}
        \includegraphics[trim={95 250 95 250}, clip=True]{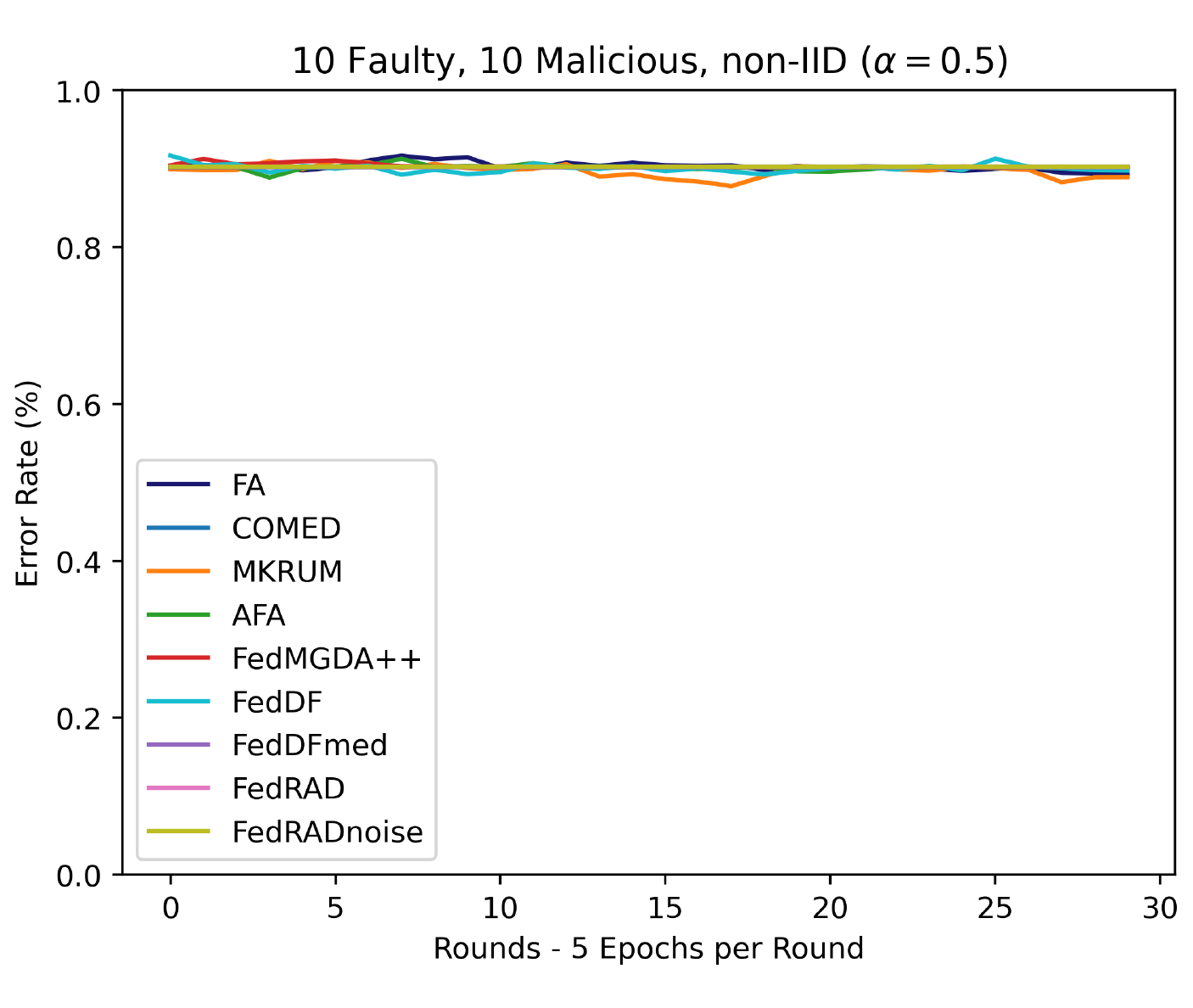}
        \includegraphics[trim={95 250 95 250}, clip=True]{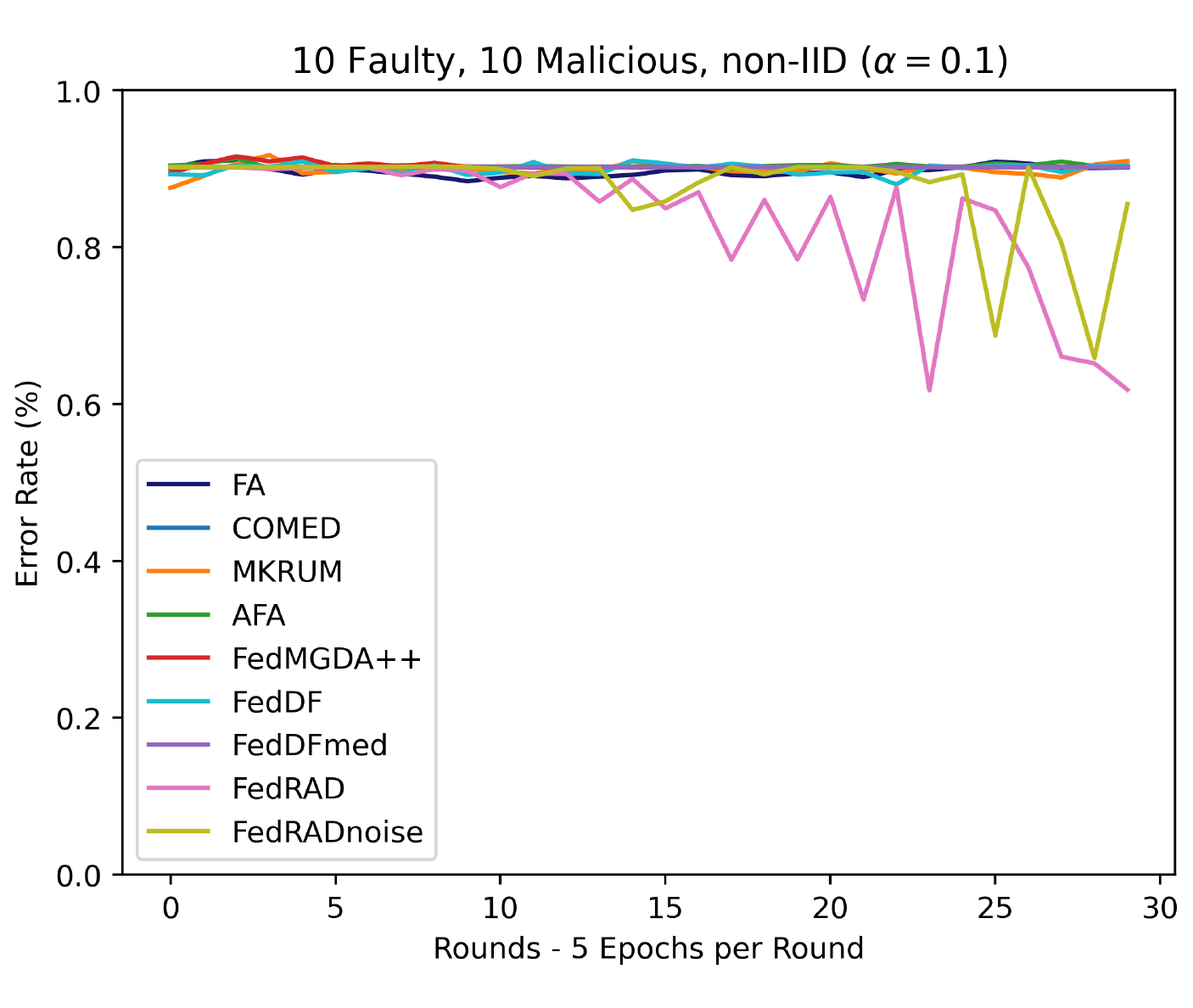}
    }
\caption{Effects of non-IID data with $10$ faulty and $10$ malicious attackers.}
\label{fig:attack_non-iid_dual10}
\end{figure}

Experiments are done using both faulty and malicious attackers in IID and non-IID settings. The results from these can be seen in Figures \ref{fig:attack_non-iid_dual5}, and \ref{fig:attack_non-iid_dual10}. 


Using 10 faulty, 10 malicious and only 10 honest agents, we finally reached the breaking point. But even with this many attakcers, our FedRAD aggregator was the only one that showed any progress and managed to reach a $60\%$ error rate on IID data, which is respectable considering only a third of the agents are honest. 

\section{Median-based knowledge distillation for different aggregators}
The median based knowledge distillation module is also tested with AFA and FedMGDA, called FedADF and FedMGDA+DF. 
FedBE and FedBEmed (which uses median-logits) were also tested.
Results are shown in Table \ref{tab:additional_results}

\begin{table}[ht]
    \centering
    \begin{tabular}{|p{2.0cm}| p{2.2cm}|>{\raggedleft\arraybackslash}p{2.4cm}|>{\raggedleft\arraybackslash}p{2.4cm}|>{\raggedleft\arraybackslash}p{2.4cm}|}
    \hline
    \multirow{3}{*}{Attacks}  & \multirow{3}{*}{Aggregator} & \multicolumn{3}{c|}{Error rates ($\%$)} \\ \cline{3-5}
& & \multicolumn{1}{c}{\multirow{2}{*}{IID}} & \multicolumn{2}{|c|}{non-IID} \\ \cline{4-5}
& & & $\alpha=0.5$ & $\alpha=0.1$ \\ \hline

\multirow{8}{2.5cm}{No attacks}
 & AFA & $\boldsymbol{5.28\pm0.17}$ & $5.94\pm0.16$ & $\boldsymbol{10.36\pm0.96}$ \\ \cline{2-5}
 & FedMGDA++ & $6.28\pm0.25$ & $12.15\pm5.39$ & $21.12\pm3.84$ \\ \cline{2-5}
 & FedBE & $\boldsymbol{5.36\pm0.13}$ & $\boldsymbol{5.93\pm0.22}$ & $11.70\pm1.38$ \\ \cline{2-5}
 & FedBEmed & $5.39\pm0.10$ & $\boldsymbol{5.85\pm0.28}$ & $\boldsymbol{9.67\pm0.29}$ \\ \cline{2-5}
 & FedADF & $5.45\pm0.14$ & $6.38\pm0.31$ & $16.81\pm4.55$ \\ \cline{2-5}
 & FedMGDA+DF & $6.27\pm0.32$ & $8.29\pm0.45$ & $15.11\pm0.57$ \\ \cline{2-5}
 & FedRAD & $\boldsymbol{5.29\pm0.16}$ & $\boldsymbol{5.82\pm0.15}$ & $\boldsymbol{9.38\pm0.57}$ \\ \cline{2-5}
 & FedRADnoise & $\boldsymbol{5.32\pm0.15}$ & $5.93\pm0.09$ & $\boldsymbol{9.71\pm0.57}$ \\ \hline \hline

\multirow{8}{2.5cm}{10 Faulty}
 & AFA & $70.97\pm32.56$ & $90.12\pm0.47$ & $89.70\pm0.33$ \\ \cline{2-5}
 & FedMGDA++ & $89.82\pm0.71$ & $89.37\pm0.28$ & $89.89\pm0.43$ \\ \cline{2-5}
 & FedBE & $90.24\pm0.85$ & $90.02\pm1.14$ & $89.87\pm0.55$ \\ \cline{2-5}
 & FedBEmed & $90.30\pm0.86$ & $90.27\pm0.72$ & $89.40\pm0.68$ \\ \cline{2-5}
 & FedADF & $\boldsymbol{5.44\pm0.18}$ & $6.45\pm0.36$ & $27.68\pm30.61$ \\ \cline{2-5}
 & FedMGDA+DF & $54.54\pm8.31$ & $90.11\pm0.32$ & $90.19\pm0.46$ \\ \cline{2-5}
 & FedRAD & $\boldsymbol{5.44\pm0.07}$ & $\boldsymbol{5.91\pm0.31}$ & $\boldsymbol{12.63\pm3.65}$ \\ \cline{2-5}
 & FedRADnoise & $\boldsymbol{5.29\pm0.21}$ & $\boldsymbol{5.99\pm0.08}$ & $\boldsymbol{13.07\pm3.29}$ \\ \hline \hline

\multirow{8}{2.5cm}{10 Malicious}
 & AFA & $17.68\pm4.46$ & $90.20\pm0.00$ & $90.20\pm0.00$ \\ \cline{2-5}
 & FedMGDA++ & $90.20\pm0.00$ & $90.20\pm0.00$ & $90.20\pm0.00$ \\ \cline{2-5}
 & FedBE & $30.40\pm3.75$ & $33.50\pm9.16$ & $29.53\pm3.10$ \\ \cline{2-5}
 & FedBEmed & $25.14\pm4.03$ & $63.77\pm32.99$ & $54.21\pm34.26$ \\ \cline{2-5}
 & FedADF & $13.11\pm2.67$ & $90.20\pm0.00$ & $84.83\pm10.74$ \\ \cline{2-5}
 & FedMGDA+DF & $70.55\pm7.44$ & $67.68\pm11.76$ & $86.42\pm5.78$ \\ \cline{2-5}
 & FedRAD & $\boldsymbol{5.89\pm0.20}$ & $\boldsymbol{6.55\pm0.43}$ & $\boldsymbol{13.09\pm1.97}$ \\ \cline{2-5}
 & FedRADnoise & $90.20\pm0.00$ & $73.45\pm33.50$ & $59.49\pm37.61$ \\ \hline \hline

\multirow{8}{2.5cm}{5 Faulty, 5 Malicious}
 & AFA & $67.62\pm6.64$ & $67.82\pm30.59$ & $80.97\pm17.03$ \\ \cline{2-5}
 & FedMGDA++ & $90.20\pm0.00$ & $90.21\pm0.00$ & $89.31\pm1.61$ \\ \cline{2-5}
 & FedBE & $90.36\pm0.28$ & $90.47\pm0.51$ & $90.25\pm0.32$ \\ \cline{2-5}
 & FedBEmed & $90.21\pm1.21$ & $90.25\pm0.42$ & $89.84\pm0.31$ \\ \cline{2-5}
 & FedADF & $6.01\pm0.19$ & $9.53\pm3.76$ & $19.47\pm8.33$ \\ \cline{2-5}
 & FedMGDA+DF & $47.88\pm4.25$ & $87.82\pm2.87$ & $89.91\pm0.82$ \\ \cline{2-5}
 & FedRAD & $\boldsymbol{5.63\pm0.20}$ & $\boldsymbol{6.15\pm0.40}$ & $\boldsymbol{10.95\pm1.22}$ \\ \cline{2-5}
 & FedRADnoise & $\boldsymbol{5.64\pm0.17}$ & $23.39\pm33.41$ & $12.84\pm1.52$ \\ 
     \hline
    \end{tabular}
    
    \caption{Test set error rate for MNIST after 30 rounds. Average and standard deviation of test-set error rates shown for $5$ different random seeds. }
\label{tab:additional_results}
\end{table}


\end{appendices}

\end{document}